\documentclass[runningheads]{llncs}
\usepackage{graphicx}

\usepackage{url}
\usepackage{hyperref}

\makeatletter
\def\orcidID#1{\smash{\href{http://orcid.org/#1}{\protect\raisebox{-1.25pt}{
\protect\includegraphics{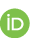}}}}}
\makeatother

\usepackage{stmaryrd}
\usepackage{listings}
\usepackage{lstautogobble}
\usepackage{capt-of}
\usepackage[T1]{fontenc}
\usepackage{textcomp}
\usepackage{amsmath}
\usepackage{xspace}
\usepackage{amsfonts}
\usepackage{amssymb}
\usepackage{enumerate}

\newcommand{\vShapeDocument}[0]{\ensuremath{M}}

\newcommand{\vn}[0]{\ensuremath{n}}
\newcommand{\vj}[0]{\ensuremath{j}}
\newcommand{\vx}[0]{\ensuremath{x}}
\newcommand{\vy}[0]{\ensuremath{y}}
\newcommand{\vz}[0]{\ensuremath{z}}
\newcommand{\vshape}[0]{\texttt{s}}
\newcommand{\vshapet}[0]{\ensuremath{t}}
\newcommand{\vshapec}[0]{\ensuremath{d}}
\newcommand{\trip}[3]{\ensuremath{\langle #1,\allowbreak #2,\allowbreak #3 \rangle}}
\newcommand{\sh}[1]{\ensuremath{\texttt{sh}{:}{\texttt{#1}}}}
\newcommand{\rdf}[1]{\ensuremath{\texttt{rdf}{:}{\texttt{#1}}}}

\newcommand{\uri}[1]{\ensuremath{{:}{\texttt{#1}}}}
\newcommand{\conc}[0]{\ensuremath{\texttt{c}}}
\newcommand{\tripsquare}[3]{\ensuremath{{<}#1,\allowbreak #2,\allowbreak #3{>}}}
\newcommand{\shapesof}[1]{\ensuremath{\mathsf{shapes}(\mathsf{#1})}}

\newcommand{\nodesof}[2]{\ensuremath{\mathsf{nodes}(\mathsf{#1},\mathsf{#2})}}
\newcommand{\GsigmaModels}[4]{\ensuremath{\llbracket #3 \rrbracket^{#4,#1,#2}}}
\newcommand{\vtrue}{\ensuremath{\mathsf{True}}}
\newcommand{\vfalse}{\ensuremath{\mathsf{False}}}
\newcommand{\vundefined}{\ensuremath{\mathsf{Undefined}}}
\newcommand{\isfaithful}[3]{\ensuremath{\ensuremath{(#1,#2) \models #3}}}

\newcommand{\minustargets}[1]{\ensuremath{#1^{\setminus t}}}

\newcommand{\assignmentsPartial}[2]{\ensuremath{A^{#1,#2}}}
\newcommand{\assignmentsTotal}[2]{\ensuremath{A_{T}^{#1,#2}}}
\newcommand{\SCL}[0]{\ensuremath{\texttt{\textbf{SCL}}}}
\newcommand{\gramdef}[0]{\ensuremath{:=}}

\newcommand{\hasshapenosubscript}[2]{\ensuremath{\Sigma(#1)}\xspace}
\newcommand{\hasshape}[2]{\ensuremath{\Sigma_{#2}(#1)}\xspace}

\newcommand{\sconst}{\texttt{s}}
\newcommand{\lO}{\texttt{O}}
\renewcommand{\S}{\texttt{S}}
\newcommand{\Z}{\texttt{Z}}
\newcommand{\A}{\texttt{A}}
\newcommand{\T}{\texttt{T}}
\newcommand{\D}{\texttt{D}}
\newcommand{\E}{\texttt{E}}

\renewcommand{\O}{\texttt{O}}
\newcommand{\C}{\texttt{C}}

\newcommand{\X}{$\varnothing$}
\newcommand{\sigmagrammar}{\ensuremath{\varsigma}}
\newcommand{\isA}[0]{\texttt{isA}\xspace}
\newcommand{\induced}[1]{\ensuremath{{\texttt{#1}^{\tau}}}}
\newcommand{\hasshapePredicateS}[1]{\ensuremath{\Sigma_{#1}}\xspace}
\newcommand{\hasshapePredicate}[0]{\ensuremath{\Sigma}\xspace}
\newcommand{\isIRI}[0]{\text{IRI}}
\newcommand{\minLength}[0]{\text{length$\geq$}}

\newcommand{\tshacl}[0]{\textsc{shacl}\xspace}
\newcommand{\tshaclplus}[0]{\textsc{shacl}$^{+}$\xspace}
\newcommand{\tshaclrec}[0]{\textsc{shacl}$^{rec}$\xspace}
\newcommand{\tshaclnonrec}[0]{\textsc{shacl}$^{non-rec}$\xspace}
\newcommand{\tsparql}[0]{\textsc{sparql}\xspace}
\newcommand{\tconstruct}[0]{\textsc{construct}\xspace}
\newcommand{\twhere}[0]{\textsc{where}\xspace}
\newcommand{\trdf}[0]{\textsc{rdf}\xspace}
\newcommand{\trdfs}[0]{\textsc{rdfs}\xspace}
\newcommand{\tiri}[0]{\textsc{iri}\xspace}
\newcommand{\tregex}[0]{\textsc{regex}\xspace}

\newcommand{\validates}[2]{\ensuremath{\textsc{Validate}(\mathsf{#1},\mathsf{#2})}}

\usepackage{tikz}
\usetikzlibrary{arrows.meta,shapes,calc,patterns}

\tikzstyle{every node} =
  [black, draw = none, thick, dashed, fill = none]

\tikzstyle{unk} =
  [brown]
\tikzstyle{dec} =
  [rounded rectangle, draw = blue, blue]
\tikzstyle{decthm} =
  [dec, very thick, solid]
\tikzstyle{und} =
  [rectangle, draw = red, red]
\tikzstyle{undthm} =
  [und, very thick, solid]
\tikzstyle{fmp} =
  [fill = blue!20!white]
\tikzstyle{notfmp} =
  [fill = red!10!white]

\tikzstyle{every edge} +=
  [black, very thin, dashed]

\tikzstyle{decder} =
  [blue, thick, solid]
\tikzstyle{decimp} =
  [decder, dotted]
\tikzstyle{undder} =https://www.overleaf.com/project/5e42733697cde1000189d483
  [red, thick, solid]

\renewcommand{\S}{\texttt{S}}
\renewcommand{\O}{\texttt{O}}

  \newcommand{\figfrg}
    {
    
    \begin{center}
    \footnotesize
    \scalebox{0.70}[0.65]
    {
    \begin{tikzpicture}
      [node distance = 4em, bend angle = 22.5]

      \node [decthm, fmp]
            (0)
            []
            {\X};

      \node [dec, fmp]
            (A)
            [above of = 0, yshift = 1em]
            {\A};
      \node [unk, notfmp]
            (O)
            [left of = A, xshift = -4em]
            {\O};
      \node [dec, fmp]
            (S)
            [left of = O, xshift = -4em]
            {\S};
      \node [dec, notfmp]
            (C)
            [right of = A, xshift = 4em]
            {\C};
      \node [dec, fmp]
            (E)
            [right of = C, xshift = 4em]
            {\E};

      \node [unk]
            (SE)
            [above of = A, yshift = 1.5em]
            {\S\,\E};
      \node [unk, notfmp]
            (AO)
            [left of = SE]
            {\A\,\O};
      \node [unk, notfmp]
            (SC)
            [left of = AO]
            {\S\,\C};
      \node [undthm, notfmp]
            (SO)
            [left of = SC]
            {\S\,\O};
      \node [dec, fmp]
            (SA)
            [left of = SO]
            {\S\,\A};
      \node [dec, notfmp]
            (AC)
            [right of = SE]
            {\A\,\C};
      \node [unk, notfmp]
            (EO)
            [right of = AC]
            {\E\,\O};
      \node [dec, notfmp]
            (EC)
            [right of = EO]
            {\E\,\C};
      \node [dec, fmp]
            (AE)
            [right of = EC]
            {\A\,\E};

      \node [unk]
            (SAE)
            [above of = SE, yshift = 1.5em]
            {\S\,\A\,\E};
      \node [und, notfmp]
            (SEO)
            [left of = SAE, xshift = -1.5em]
            {\S\,\E\,\O};
      \node [undthm, notfmp]
            (SAC)
            [left of = SEO]
            {\S\,\A\,\C};
      \node [und, notfmp]
            (SAO)
            [left of = SAC]
            {\S\,\A\,\O};
      \node [undthm, notfmp]
            (SEC)
            [right of = SAE, xshift = 1.5em]
            {\S\,\E\,\C};
      \node [unk, notfmp]
            (AEO)
            [right of = SEC]
            {\A\,\E\,\O};
      \node [dec, notfmp]
            (AEC)
            [right of = AEO]
            {\A\,\E\,\C};

      \node [undthm, notfmp]
            (SZAE)
            [above of = SAE, xshift = -2.5em]
            {\S\,\Z\,\A\,\E};
      \node [decthm, fmp]
            (SZAD)
            [above of = SA, yshift = 5.5em, xshift = -1.5em]
            {\S\,\Z\,\A\,\D};
      \node [unk]
            (SADE)
            [above of = SAE, xshift = 2.5em]
            {\S\,\A\,\D\,\E};
      \node [decthm, fmp]
            (ZADE)
            [above of = AE, yshift = 5.5em, xshift = 1.5em]
            {\Z\,\A\,\D\,\E};

      \node [und, notfmp]
            (SZADE)
            [node distance = 9em, above of = SAE]
            {\S\,\Z\,\A\,\D\,\E};
      \node [und, notfmp]
            (SAEO)
            [left of = SZADE, xshift = -2em]
            {\S\,\A\,\E\,\O};
      \node [und, notfmp]
            (SZADC)
            [left of = SAEO, xshift = -1em]
            {\S\,\Z\,\A\,\D\,\C};
      \node [und, notfmp]
            (SZADO)
            [left of = SZADC, xshift = -1em]
            {\S\,\Z\,\A\,\D\,\O};
      \node [und, notfmp]
            (SAEC)
            [right of = SZADE, xshift = 2em]
            {\S\,\A\,\E\,\C};
      \node [unk, notfmp]
            (ZADEO)
            [right of = SAEC, xshift = 1em]
            {\Z\,\A\,\D\,\E\,\O};
      \node [decthm, notfmp]
            (ZADEC)
            [right of = ZADEO, xshift = 1em]
            {\Z\,\A\,\D\,\E\,\C};

      \node [und, notfmp]
            (SZATE)
            [above of = SZADE, xshift = -2.5em]
            {\S\,\Z\,\A\,\T\,\E};
      \node [decthm, notfmp]
            (SZATD)
            [above of = SZAD, yshift = 5em]
            {\S\,\Z\,\A\,\T\,\D};
      \node [unk]
            (SATDE)
            [above of = SZADE, xshift = 2.5em]
            {\S\,\A\,\T\,\D\,\E};
      \node [unk]
            (ZATDE)
            [above of = ZADE, yshift = 5em]
            {\Z\,\A\,\T\,\D\,\E};

      \node [und, notfmp]
            (SZATDE)
            [node distance = 8em, above of = SZADE]
            {\S\,\Z\,\A\,\T\,\D\,\E};
      \node [und, notfmp]
            (SZATDO)
            [above of = SZATD]
            {\S\,\Z\,\A\,\T\,\D\,\O};
      \node [und, notfmp]
            (SZATDC)
            [node distance = 6em, right of = SZATDO]
            {\S\,\Z\,\A\,\T\,\D\,\C};
      \node [unk, notfmp]
            (ZATDEC)
            [above of = ZATDE]
            {\Z\,\A\,\T\,\D\,\E\,\C};
      \node [unk, notfmp]
            (ZATDEO)
            [node distance = 6em, left of = ZATDEC]
            {\Z\,\A\,\T\,\D\,\E\,\O};

      \node [und, notfmp]
            (SZATDEO)
            [above of = SAEO, yshift = 8em]
            {\S\,\Z\,\A\,\T\,\D\,\E\,\O};
      \node [und, notfmp]
            (SZATDOC)
            [above of = SZATDO, xshift = 3em]
            {\S\,\Z\,\A\,\T\,\D\,\O\,\C};
      \node [und, notfmp]
            (SZATDEC)
            [above of = SAEC, yshift = 8em]
            {\S\,\Z\,\A\,\T\,\D\,\E\,\C};
      \node [unk, notfmp]
            (ZATDEOC)
            [above of = ZATDEC, xshift = -3em]
            {\Z\,\A\,\T\,\D\,\E\,\O\,\C};

      \path[Latex-Latex]
        (0)     edge  [decder]
                      (A)
        (0.west)
                edge  [decder]
                      (S)
        ;
      \path[-Latex]
        (0)     edge  []
                      (O)
                edge  [decimp]
                      (C)
        (0.east)
                edge  [decimp]
                      (E)
        ;

      \path[Latex-Latex]
        (A)     edge  [decder]
                      (SA.south)
        (O.north)
                edge  []
                      (AO.south)
        (S.north)
                edge  [decder]
                      (SA.south)
        ;
      \path[-Latex]
        (A.north)
                edge  []
                      (AO.south)
                edge  [decimp]
                      (AC.south)
        (A)     edge  [decimp]
                      (AE.south)
        (O.north)
                edge  []
                      (SO.south)
                edge  []
                      (EO.south)
        (S.north)
                edge  []
                      (SE.south)
                edge  []
                      (SC.south)
                edge  []
                      (SO.south)
                edge  [decder]
                      (SA.south)
        (C.north)
                edge  []
                      (SC.south)
                edge  [decder]
                      (AC.south)
                edge  [decder]
                      (EC.south)
        (E.north)
                edge  []
                      (SE.south)
                edge  []
                      (EO.south)
                edge  [decimp]
                      (EC.south)
                edge  [decder]
                      (AE.south)
        ;

      \path[-Latex]
        (SE.north)
                edge  []
                      (SAE.south)
                edge  []
                      (SEC.south)
                edge  []
                      (SEO.south)
        (AO.north)
                edge  []
                      (SAO.south)
                edge  []
                      (AEO.south)
        (SC.north)
                edge  []
                      (SAC.south)
                edge  []
                      (SEC.south west)
        (SO.north)
                edge  [undder]
                      (SAO.south)
                edge  [undder]
                      (SEO.south)
        (SA.north)
                edge  []
                      (SAE.south)
                edge  []
                      (SAC.south)
                edge  []
                      (SAO.south)
        (AC.north)
                edge  []
                      (SAC.south)
                edge  [decder]
                      (AEC.south)
        (EO.north)
                edge  []
                      (SEO.south east)
                edge  []
                      (AEO.south)
        (EC.north)
                edge  []
                      (SEC.south)
                edge  [decder]
                      (AEC.south)
        (AE.north)
                edge  []
                      (SAE.south)
                edge  [decimp]
                      (AEC.south)
                edge  []
                      (AEO.south)
        ;

      \path[-Latex]
        (SA.north)
                edge  [decimp]
                      (SZAD.south)
        (AE.north)
                edge  [decder]
                      (ZADE.south)
        ;

      \path[-Latex]
        (SAE)   edge  []
                      (SZAE)
                edge  []
                      (SADE)
        ;

      \path[-Latex]
        (SAE)   edge  [bend angle = 40, bend left]
                      (SAEO.south east)
                edge  [bend angle = 40, bend right]
                      (SAEC.south west)
        (SEO)   edge  [undder]
                      (SAEO.south)
        (SAC.north)
                edge  [undder]
                      (SZADC.south)
                edge  [undder, bend left]
                      (SAEC.south west)
        (SAO.north)
                edge  [undder]
                      (SZADO.south)
                edge  [undder]
                      (SAEO.south)
        (SEC)   edge  [undder]
                      (SAEC.south)
        (AEO.north)
                edge  []
                      (ZADEO.south)
                edge  [bend right]
                      (SAEO.south east)
        (AEC.north)
                edge  [decder]
                      (ZADEC.south)
                edge  []
                      (SAEC.south)
        ;

      \path[-Latex]
        (SZAE.north)
                edge  [undder]
                      (SZADE.south)
        (SZAD)  edge  []
                      (SZADE.south)
        (SZAD.north)
                edge  []
                      (SZADO.south)
                edge  []
                      (SZADC.south)
        (SADE.north)
                edge  []
                      (SZADE.south)
        (ZADE)  edge  []
                      (SZADE.south)
        (ZADE.north)
                edge  []
                      (ZADEO.south)
                edge  [decimp]
                      (ZADEC.south)
        ;

      \path[-Latex]
        (SZAE.north)
                edge  [undder]
                      (SZATE.south)
        (SZAD.north)
                edge  [decimp, bend left]
                      (SZATD.south)
        (SADE.north)
                edge  []
                      (SATDE.south)
        (ZADE.north)
                edge  [bend right]
                      (ZATDE.south)
        ;

      \path[-Latex]
        (SZADE) edge  [undder]
                      (SZATDE)
        (SZADC) edge  [undder]
                      (SZATDC.south)
        (SZADO) edge  [undder, bend angle = 35, bend right]
                      (SZATDO)
        (ZADEO) edge  []
                      (ZATDEO.south)
        (ZADEC) edge  [bend angle = 35, bend left]
                      (ZATDEC)
        ;

      \path[-Latex]
        (SAEO)  edge  [undder]
                      (SZATDEO.south)
        (SAEC)  edge  [undder]
                      (SZATDEC.south)
        ;

      \path[-Latex]
        (SZATE) edge  [undder]
                      (SZATDE)
        (SZATD) edge  []
                      (SZATDE)
        (SZATD.north)
                edge  []
                      (SZATDO.south)
                edge  []
                      (SZATDC.south)
        (SATDE) edge  []
                      (SZATDE)
        (ZATDE) edge  []
                      (SZATDE)
        (ZATDE.north)
                edge  []
                      (ZATDEO.south)
                edge  []
                      (ZATDEC.south)
        ;

      \path[-Latex]
        (SZATDE.north)
                edge  [undder]
                      (SZATDEO.south)
                edge  [undder]
                      (SZATDEC.south)
        (SZATDC.north)
                edge  [undder]
                      (SZATDEC.south west)
                edge  [undder]
                      (SZATDOC.south)
        (SZATDO.north)
                edge  [undder]
                      (SZATDEO.south)
                edge  [undder]
                      (SZATDOC.south)
        (ZATDEO.north)
                edge  []
                      (SZATDEO.south east)
                edge  []
                      (ZATDEOC.south)
        (ZATDEC.north)
                edge  []
                      (SZATDEC.south)
                edge  []
                      (ZATDEOC.south)
        ;

    \end{tikzpicture}
    }
    \end{center}
    }

\begin{document}
\title{A Review of SHACL: From Data Validation to Schema Reasoning for RDF Graphs}
\titlerunning{A Review of \tshacl}
\author{Paolo Pareti\inst{1}\orcidID{0000-0002-2502-0011} \and
George Konstantinidis\inst{2}\orcidID{0000-0002-3962-9303} }
\authorrunning{Pareti and Konstantinidis}
%
\institute{University of Winchester, Winchester, United Kingdom 
\email{paolo.pareti@winchester.ac.uk}\\
\and
University of Southampton, Southampton, United Kingdom\\
\email{g.konstantinidis@soton.ac.uk}}
\maketitle              
%
\begin{abstract}

We present an introduction and a review of \emph{Shapes Constraint Language} (\tshacl), the W3C recommendation language for validating \trdf data. A \tshacl document describes a set of constraints on \trdf nodes, and a graph is valid with respect to the document if its nodes satisfy these constraints. We revisit the basic concepts of the language, its constructs and components and their interaction. We review the different formal frameworks used to study this language and the different semantics proposed. We examine a number of related problems, from containment and satisfiability to the interaction of \tshacl with inference rules, and exhibit how different modellings of the language are useful for different problems. We also cover practical aspects of \tshacl, discussing its implementations and state of adoption, to present a holistic review useful to practitioners and theoreticians alike.
\end{abstract}

\section{Introduction}

The Shapes Constraint Language (\tshacl) \cite{2017SHACL} is a W3C recommendation language for the validation of RDF graphs. In \tshacl, validation is based on \emph{shapes}, which define particular constraints and specify which nodes in a graph should be validated against these constraints. The ability to validate data with respect to a set of constraints is of particular importance for RDF graphs, as they are schemaless by design. Validation can be used to detect problems in a dataset and it can provide data quality guarantees for the purpose of data exchange and interoperability. A set of constraints can also be interpreted as a ``schema'', functioning as one of the primary descriptors of a graph dataset, thus enhancing its understandability and usability. A set of \tshacl shapes is called a \emph{shapes graph}, but we refer to it as a \tshacl \emph{document} in order not to confuse it with the graphs that it is used to validate.

In this article we present a review of \tshacl, which is composed of three main parts. In the first part, in Sections \ref{sec:preliminaries} and \ref{sec:shacloverview}, we review the \tshacl specification. This part focuses on how shapes are defined, and how they are used for the purpose of validation. We highlight the main peculiarities of this language, and discuss how \tshacl validation can be expressed either in terms of \tsparql queries, to facilitate its implementation, or in terms of \emph{assignments} \cite{Corman2018SHACL}, to make it amenable to theoretical study. The syntax of \tshacl is outside the scope of this review, and for the precise details on how to encode particular constraints we refer the reader to the \tshacl specification \cite{2017SHACL}. We also do not discuss the process that lead to the development of \tshacl, but it should be noted that this specification was built on top of a number of previous constraint languages, the most influential of which is Shape Expressions (ShEx) \cite{ShEX2019}.

In the second part of our review, in Sections \ref{sec:recursion} to \ref{sec:schemaexpansion}, we present the formal properties of this language. This mainly revolves around a discussion of \emph{recursion}.
The semantics of recursion is not defined in the \tshacl specification, and thus has been the subject of significant subsequent research \cite{SHACLstableModelSemantics}. The formal semantics of \tshacl is given as a translation into \SCL\ \cite{pareti2020}, a first order logic language that captures the entirety of the \tshacl specification. Apart from validation, several standard decision problems are discussed, such as \emph{satisfiability} and \emph{containment}, along with an existing study on the interaction of \tshacl with inference rules. We try to keep a consistent notation throughout this article and at times this notation might be different from the one in the original articles.

In Section \ref{sec:relatedwork}, we review existing implementations of \tshacl validators and their integration with mainstream graph databases. We also review prominent additional tools to manage \tshacl documents, such as tools designed to automate or semi-automate the process of creating \tshacl documents by exploiting graph data, ontologies, or other constraint languages. These approaches provide solutions to the cold start problem, and alleviate reliance on expert knowledge, which are typical problems of new technologies. We complement a discussion of these approaches with a review of prominent applications of \tshacl in several domains; in summary, the abundance of \tshacl related tools and applications highlights the remarkable level of maturity and adoption reached by this relatively new language.

\section{Preliminaries} \label{sec:preliminaries}

Before discussing \tshacl, we briefly introduce our notation for \trdf graphs \cite{Hayes2014GeneralisedRDF}. With the term \emph{\trdf graph} (or just \emph{graph}) we refer to a set of \trdf \emph{triples} (or just \emph{triples}), where each triple \tripsquare{s}{p}{o} identifies an edge with label $p$, called \emph{predicate}, from a node $s$, called \emph{subject}, to a node $o$, called \emph{object}. Subjects, predicates and objects of \trdf triples are collectively called \trdf \emph{terms}. The \trdf terms that appear as subject and objects in the triples of a graph are called the \emph{nodes} of the graph. Graphs in this article are represented in Turtle syntax using common XML namespaces, such as \texttt{sh}, \texttt{rdf} and \texttt{rdfs} to refer to, respectively, the \tshacl, \trdf, and \trdfs \cite{RDFS} vocabularies. Queries over \trdf graph will be expressed as \tsparql \cite{SeaborneSPARQL2013} queries.

In the \trdf data model, subjects, predicates and objects are defined over different but overlapping domains. For example, while \trdf terms of the \tiri type can occupy any position in a triple, \trdf terms of the \emph{literal} type (representing datatype values) can only appear in the object position. These differences are not central to the topics discussed in this review, and thus, for the sake of simplifying notation, we will assume that all elements of a triple are drawn from a single and infinite domain of constants. This corresponds to the notion of \emph{generalized} \trdf \cite{Hayes2014GeneralisedRDF}.

\section{Overview of \tshacl}\label{sec:shacloverview}

\begin{figure}[t]
\begin{minipage}[t]{0.60\linewidth}
\begin{lstlisting}
:EmployeeShape a sh:PropertyShape ;
    sh:targetClass :Employee ;
    sh:path :hasOfficeNumber ;
    sh:minCount 1 .
\end{lstlisting}
\end{minipage}\hfill
\begin{minipage}[t]{0.40\linewidth}
\begin{lstlisting}
:Anne a :Employee .
:Bob a :Employee ;
    :hasOfficeNumber "18" ;
    :hasOfficeNumber "3" .
:Carl a :Employee ;
    :hasOfficeNumber "171" .
:David a :Customer .
\end{lstlisting}
\end{minipage}
\captionof{figure}{(Left) A sample \tshacl document (shape graph) stating the constraint that every employee must have at least one office number. (Right) A sample \trdf graph (data graph).}\label{fig:exampleSHACL1}
\end{figure}

The main application of \tshacl is data validation. Data validation in \tshacl requires two inputs: (1) an \trdf graph $G$ to be validated and (2) a \tshacl document $\vShapeDocument$ that defines the conditions against which $G$ must be evaluated. The \tshacl specification defines the output of the data validation process as a \emph{validation report}, detailing all the violations that were found in $G$ of the conditions set by $\vShapeDocument$. If the violation report contains no violations, a graph $G$ is \emph{valid} w.r.t.\ \tshacl document $\vShapeDocument$. The \tshacl validation process can be abstracted into the following decision problem. Given a graph $G$ and a \tshacl document $\vShapeDocument$, we denote with \validates{G}{\vShapeDocument} the decision problem of deciding whether $G$ is \emph{valid} w.r.t.\ \tshacl document $\vShapeDocument$, that we call \emph{validating} $G$ against $\vShapeDocument$.

For example, the graph on the left of Figure \ref{fig:exampleSHACL1} represents a \tshacl document $\vShapeDocument_{1}$, that defines the condition that every employee must have an office number. Therefore, the validation report for a graph and $\vShapeDocument_{1}$ would list all of the instances of \uri{Employee} in the graph that do not have an office number. The validation report for $\vShapeDocument_{1}$ and the data graph $G_{1}$ on the right of Figure \ref{fig:exampleSHACL1} contains a violation on node \rdf{Anne}, since she does not have an office number. Therefore $G_{1}$ is not valid w.r.t. $\vShapeDocument_{1}$.

Formally, a \tshacl \emph{document} is a set of \emph{shapes}. Validating a graph against a \tshacl document involves validating it against each shape. Shapes restrict the structure of a valid graph by focusing on certain nodes and examining whether they satisfy their constraints. The main components of a shape are a \emph{constraint} $\vshapec$ and a \emph{target definition} $\vshapet$. Constraints can be \emph{evaluated} on any \trdf node to determine whether that node \emph{satisfies} or not the given constraints. A node that satisfies the constraint of a shape it is said to \emph{conform} to that shape, or \emph{not-conform} otherwise. If a shape has an empty constraint, all nodes trivially conform to the shape. 
Not all nodes of a graph must conform to all the shapes in the \tshacl document. The constraint definition of each shape defines which \trdf nodes, called \emph{target nodes}, must conform to that shape in order for the graph to be valid. A shape with an empty constraint definition does not have any target nodes. Through inter-shape referencing, as we will see below, additional nodes might be required to conform to certain shapes (or not, if negation is used) for the validation to succeed. Further irrelevant nodes within the graph do not play a role in validation of the shape, whether they conform to it or not. The \tshacl document $\vShapeDocument_{1}$ of our previous example, contains shape \uri{EmployeeShape}, whose constraint captures the property of ``having an office number'', and whose target definition targets only the \trdf nodes of type Employee. Nodes of the Client type do not generate violations by not having an office number.  

Formally, a shape is a tuple \trip{\vshape}{\vshapet}{\vshapec} defined by three components: (1) the shape name $\vshape$, which uniquely identifies the shape; (2) the \emph{target definition} $\vshapet$, and (3) the set of \emph{constraints} which are used in conjunction, and hence hereafter referred to as the single constraint $\vshapec$.
As demonstrated in Figure \ref{fig:exampleSHACL1}, a \tshacl document is itself an \trdf graph. The graph representing a \tshacl document is called a \emph{shapes graph}, while the graph being validated is called a \emph{data graph}. This approach to serialisation is similar to how OWL ontologies are serialised, and it serves a similar purpose. Thanks to this approach, a \tshacl document does not require any dedicated infrastructure to be stored and shared. In fact, a \tshacl document can be embedded directly into the very graph it validates, thus combining the shape graph and data graph into a single graph. 
Interestingly, with this serialisation, a shapes graph, being an \trdf graph, can be itself subject to validation. The \tshacl specification, in fact, defines a shapes graph that can be used to validate shapes graphs.

We will now look in more details at the two major components of shapes, namely target definitions and constraints.

\subsection{\tshacl Target Definitions}

A \tshacl target definition, within a constraint, is a set of \emph{target declarations}. There are four types of target declarations defined in \tshacl, each one taking an \trdf term $\conc$ as a parameter.
\begin{description}
    \item[Node Targets] A node target declaration on $\conc$ targets that specific node.
    \item[Class-based Targets] If a shape has a class-based target on $\conc$, then all the nodes in the graph that are of type (\rdf{type}) $\conc$ are target nodes for that shape.
    \item[Subjects-of Targets] If a shape has a subject-of target on $\conc$, then the target nodes for that shape are all the nodes in the graph that appear as subjects in triples with $\conc$ as the predicate.
    \item[Objects-of Targets] If a shape has an object-of target on $\conc$, then the target nodes for that shape are all the nodes in the graph that appear as objects in triples with $\conc$ as the predicate.
\end{description}
The shape defined in Figure \ref{fig:exampleSHACL1} demonstrates an example of a class-based target targeting class \uri{Employee}. Similarly, to target a subject of a property, e.g., \uri{worksAt}, the second line of the shape definition would be substituted with: \begin{verbatim}
    sh:targetSubjectOf :worksAt.
\end{verbatim} 

Typically, a target declaration is used to select, among all the nodes in a graph, the ones to target for constraint validation. The \emph{node target} declaration, however, behaves differently, as it targets a particular node regardless of whether this node occurs in the graph or not. An important implication of this is that empty graphs are not trivially valid, since node targets can detect violations on nodes external to the graph. 
If a target definition of a shape is empty, then that shape will have no target nodes. However, this does not mean that the constraint of that shape will not be evaluated on any nodes since, as mentioned, other shapes can refer to it and ``pass it'' a node to check for conformance. 

\subsection{Focus nodes and property paths}

When a target or another node is considered against a shape for conformity, we call it a \emph{focus} node. Initially a shape focuses on its target nodes (these are the initial set of focus nodes). Additional focus nodes are obtained by following \emph{\tshacl property paths}, which we also refer to as just \emph{paths}. \tshacl property paths are a subset of \tsparql \emph{property paths} and, as the name suggests, define paths in the \trdf graph. The simplest type of path, called predicate path, corresponds to a single property IRI $\conc$. This path identifies all the nodes that are reachable in the \trdf graph from the current focus node by following a single edge $\conc$. In other words, this path identifies all the \trdf nodes in the object position of triples that have $\conc$ as the predicate and the current focus node as the subject. More complex paths can be constructed by inverting the direction of a path, by concatenating two different paths one after the other, or by allowing the repetition of a path for a minimum, maximum or arbitrary number of times.

Based on the use of property paths, \tshacl specification distinguishes shapes into two types: \emph{node shapes} and \emph{property shapes}. Intuitively, the constraint of a node shape is evaluated directly on the focus nodes of the shape. Instead, when using a property path, shapes must be declared as a property shapes. These are characterised by a path, and their constraints are evaluated over all of the nodes that can be reached from the focus nodes following such path. For example, the constraint that every employee's password must be at least 8 characters long can be represented by a property shape that targets employee nodes, and that has a relation such as \uri{hasPassword} as its path. In this way, the actual nodes that must satisfy the ``at least 8 characters long'' constraint are not the target nodes, but instead those that appear as objects in triples with an employee node as a subject, and \uri{hasPassword} as the predicate.

\subsection{\tshacl Constraints}

The majority of the \tshacl recommendation is dedicated to defining the different types of constraint components that can be used in \tshacl constraints. The main type of constraint components are called \emph{core constraint components}. These are the components that \tshacl compliant systems typically support, and where most of the existing literature focuses on. The other main type of components are the \emph{\tsparql-based constraint components}, that are used to embed \tsparql queries into \tshacl constraints. This significantly increases the expressive power of such constraints. However, the inclusion of arbitrarily complex \tsparql queries can lead to performance issues, and can make such constraints harder to understand and use. It is also worth noting that, outside of the \tshacl recommendation,  a number of additional \tshacl features\footnote{\url{https://w3c.github.io/shacl/shacl-af/} accessed on 18/6/21} are currently being designed, and some of them might be included in further versions of \tshacl. In the rest of this paper we will focus on core constraint components.

In order to better understand \tshacl core constraint components, we propose a broad categorisation of these components into three main categories, depending on how they are evaluated on the focus nodes. Notice that most constraint components can be used in both node shapes and property shapes. 

\begin{description}
    \item[Graph Structure Components.] These components define constraints that are evaluated at the level of triples of the graph, and focus on restrictions such as the minimum and maximum cardinality that the focus node must have for certain paths, or the \trdf class that the focus node should be a type of. The shape defined in Figure \ref{fig:exampleSHACL1} demonstrates an example of a minimum cardinality constraint for predicate path \uri{hasOfficeNumber}. 
    Two other salient constraints in this category are the property pair equality and disjointedness, that specify whether the two sets of nodes reachable from two different paths must be equal or disjoint, respectively.
    \item[Filter Components.] These components define constraints that are evaluated at the level of nodes, and their evaluation is usually independent from the triples present in the graph. Filter constraints restrict the focus node (1) to be a particular \trdf term, (2) to be of a particular type, such as IRI, blank node or literal, or (3) to be a literal that satisfies certain properties, such as being of the integer datatype, or a string produced by a certain regular expression. 
    \item[Logical Components.] Logical components define the standard logical operators of conjunction, disjunction and negation over other constraints. 
\end{description}
While most core constraint components fall into one of these categories, the pair of constraints \sh{lessThan} and \sh{lessThanOrEquals} is a notable exception, as it is combines the properties of graph structure and filter components. These two constraints require all the nodes reachable by one path to be literals that are less than (resp. less than or equals) to the nodes reachable by a second path.

It is worth noting that all constraints but one, namely \sh{closed}, are not affected by triples with unknown predicates (i.e.\ predicates not occurring in the \tshacl document). This means that if a graph is valid with respect to a set of those constraints, it would still remain valid if new triples with unknown predicates are added to the graph. Thus, given a non-empty graph $G$, valid w.r.t.\ a SHACL document $\vShapeDocument$,  graph $G \, \cup \, \tripsquare{s}{p}{o}$ is also valid w.r.t.\ $\vShapeDocument$ if (1) $p$ does not occur in $\vShapeDocument$ and (2) $\vShapeDocument$ does not contain the \sh{closed} constraint component. Intuitively, this means that those constraints restrict the usage of terms from a particular vocabulary, but they do not restrict in any way the graph from containing triples described using other vocabularies. The \sh{closed} component, on the other hand, restricts the predicates of the triples that have the focus node as a subject to belong to a predetermined finite set. Effectively, the \sh{closed} component can prohibit the use of unknown predicate relations for certain nodes in the graph, and thus prevent the inclusion of terms from other vocabularies. Interestingly, component \sh{closed} introduces an asymmety in \tshacl, since it only affects triples where the focus node is the subject, and it is not possible to define a similar constraint for nodes in the object position.

A major feature of \tshacl is that constraints can use the name of a shape to require a particular set of nodes to conform to that shape. This is called a \emph{shape reference}. An example of a shape reference is demonstrated by the \tshacl document in Figure \ref{fig:exampleSHACL2}. This document contains shape \uri{EmployeeShapeB} which references shape \uri{OfficeNumberShape}. The former shape restricts all of its target nodes to having an edge \uri{hasOfficeNumber} to a node that conforms to the latter shape, having a string length of at least three characters. Validating the data graph in Figure \ref{fig:exampleSHACL1} with the \tshacl document in Figure \ref{fig:exampleSHACL2} results in two violating nodes for shape \uri{EmployeeShapeB}. The first one is \uri{Anne}, who does not have an office number, and the second one is \uri{Bob}, whose office numbers all contain fewer than three digits.

\begin{figure}[t]
\begin{center}
\begin{tabular}{c}
\begin{lstlisting}
:EmployeeShapeB a sh:PropertyShape ;
   sh:targetClass :Employee ;
   sh:path :hasOfficeNumber ;
   sh:qualifiedMinCount 1 ;
   sh:qualifiedValueShape :OfficeNumberShape .

:OfficeNumberShape a sh:NodeShape ;
   sh:minLength 3 .
\end{lstlisting}
\end{tabular}
\end{center}
\captionof{figure}{A sample \tshacl document stating the constraint that every employee must have at least one 3-characters or longer office number.}\label{fig:exampleSHACL2}
\end{figure}

Shape references can be recursive, that is, the constraint of a shape can reference the constraints of a second shape which, in turn, can reference the constraints of a third shape, and so on, creating a loop. 
Let $S_{0}^{\vshapec}$ be the set of all the shape names occurring in a constraint $\vshapec$ of a shape $\trip{\vshape}{\vshapet}{\vshapec}$; these are the \emph{directly} referenced shapes of $\vshape$.
Let $S_{i+1}^{\vshapec}$ be the set of shapes in  $S_{i}^{\vshapec}$ union the directly referenced shapes of the constraints of the shapes in $S_{i}^{\vshapec}$ . 
\begin{definition}
A shape $\trip{\vshape}{\vshapet}{\vshapec}$ is \emph{recursive} if $\vshape \in S_{\infty}^{\vshapec}$; else it is \emph{non-recursive}.
\end{definition}
\begin{definition}
A \tshacl document $\vShapeDocument$ is \emph{recursive} if it contains a recursive shape, and \emph{non-recursive} otherwise. 
\end{definition}
The semantics of recursive \tshacl documents are not defined in the \tshacl specification. In Section \ref{sec:validation} we review the official semantics of non-recursive \tshacl documents, while in Section \ref{sec:recursion} we review the extended semantics for recursive \tshacl document that have been proposed in the literature.

\section{\tshacl Validation}\label{sec:validation}

In this section we present the semantics of \tshacl data validation, that is, the \validates{G}{\vShapeDocument} decision problem, for any given graph $G$ and \tshacl document $\vShapeDocument$.
In Section \ref{sec:validationBySPARQL} we review how validation is defined in the  \tshacl specification, with the help of \tsparql queries. While this query-based description of  \tshacl semantics can be easily translated into a concrete implementation, it does not lend itself well to theoretical investigation. In Section \ref{sec:validationByAssignments} we will discuss an alternative approach to defining  \tshacl semantics that is instead amenable to a formal study.

\subsection{\tshacl Validation by \tsparql queries}\label{sec:validationBySPARQL}

The validation of an \trdf graph $G$ against a \tshacl document $\vShapeDocument$ can be performed on a shape-by-shape basis. For each shape $\trip{\vshape}{\vshapet}{\vshapec}$, this process involves verifying the fact that every node $\vn$, targeted by target definition $\vshapet$, satisfies constraint $\vshapec$. 
Intuitively, graph $G$ is valid w.r.t.\ $\vShapeDocument$ if and only if this fact is true for every shape in $\vShapeDocument$.

Given a graph $G$ and a target definition $\vshapet$, the set of target nodes for $\vshapet$ can be computed by evaluating a \tsparql query on $G$ for each target declaration in $\vshapet$, and taking the union of the values returned by these queries. Table \ref{tab:translationTargsToSPARQL} details the corresponding \tsparql query for each of the four types of target declarations defined in \tshacl. It should be noted that, by default, \tshacl does not enforce any particular entailment regime. If an entailment regime is being adopted, then this should be taken into account when developing a \tshacl validator. For example, if the \trdfs entailment regime \cite{RDFS} is being considered, subclass inference should be accounted for when computing the set of entities of a given class. To accommodate for this entailment regime, the query for the node target in Table \ref{tab:translationTargsToSPARQL} could be updated to the following one.
\begin{lstlisting}[escapeinside={(*}{*)}]
SELECT ?x WHERE {  
  ?x rdf:type/rdfs:subClassOf* (*$\conc$*) 
}
\end{lstlisting}

\begin{table}[t]
\begin{center}
\caption{Target declarations and their corresponding \tsparql queries to compute the set of target nodes on a given graph}
\label{tab:translationTargsToSPARQL}
\begin{tabular}{ |l | l |}
 \hline
 Target declaration & \tsparql query \\ \hline 
 Node target (node \conc) & SELECT ?x WHERE \{ VALUES ?x \{ $\conc$ \} \} \\ \hline
 Class target (class \texttt{c})  & SELECT ?x WHERE \{ ?x $\rdf{type}$ $\conc$ . \}
    \\ \hline
 Subjects-of target (predicate $\conc$)  & SELECT ?x WHERE \{ ?x $\conc$ ?y . \} \\ \hline
 Objects-of target (relation $\conc$)  & SELECT ?x WHERE \{ ?y $\conc$ ?x . \} \\ \hline
\end{tabular}
\end{center}
\end{table}

Once an \trdf term has been identified as being in the target of a shape, evaluating whether it conforms to the shape can be done using \tsparql queries. In the \tshacl specification, in fact, several core constraint components are defined with respect to \tsparql queries. Most notably, the semantics of \tshacl filter components is in direct dependence to the semantics of \tsparql filter functions. For example, the \sh{minLength} constraint component restricts a focus node to having a string length equal or larger than a given number. Formally, a focus node $\vn$ has a \sh{minLength} of $\vj$ if and only if the following \tsparql query evaluates to true.
\begin{lstlisting}[escapeinside={(*}{*)}]
ASK { 
  FILTER (STRLEN(str((*$\vn$*))) >= (*$\vj$*)) . 
}
\end{lstlisting}

Not all \tshacl constraints, however, can be easily verified by a single \tsparql query. Evaluating whether a constraint that contains shape references is satisfied by a focus node, in fact, might involve evaluating whether other constraints are satisfied by other nodes which, in turn, might require even further constraint evaluations. For example, in order to evaluate whether node \rdf{Carl} from the data graph in Figure \ref{fig:exampleSHACL1} conforms to shape \rdf{EmployeeShapeB} from Figure \ref{fig:exampleSHACL2}, we would need to evaluate whether his office number, namely \trdf term ``171'', conforms to shape \rdf{OfficeNumberShape}. This is especially problematic in case of recursion, as it could generate an infinite series of constraint evaluations. 
For non-recursive \tshacl documents, Corman et al.\ \cite{SHACL2SPARQLtranslation} showed that it is always possible to check the validity of a graph using a single \tsparql query. For example, a graph can be checked against the \tshacl document of Figure \ref{fig:exampleSHACL2} by evaluating the following \tsparql query. 
\begin{lstlisting}[escapeinside={(*}{*)}]
SELECT ?x WHERE { 
  ?x a :Employee . 
  FILTER NOT EXISTS {
    ?x :hasOfficeNumber ?y . 
    FILTER (STRLEN(str(?y)) >= 3) .
  }
}
\end{lstlisting}
This query selects all \trdf nodes of type Employee that do not have an office number with at least three characters. Thus, any \trdf term returned by this query is a node violating a shape of the \tshacl document. If this query evaluates to an empty set, then the graph that it is evaluated on is valid with respect to the \tshacl document.

\subsection{Shape Assignments: A Tool for Defining \tshacl Validation}\label{sec:validationByAssignments}

The \tsparql-based approach to \tshacl validation does not provide a concise and formal description of \tshacl semantics. Moreover, it does not provide us with a terminating procedure to check graphs in the face of \tshacl recursion. In this section we review the concept of \emph{shape assignments} (or just  \emph{assignments}) \cite{Corman2018SHACL}, which can be used to address the above mentioned problems.

As defined in Table \ref{tab:translationTargsToSPARQL}, a target declaration $\vshapet$ is a unary query over a graph $G$. We denote with $G \models \vshapet(\vn)$ that a node $\vn$ is \emph{in the target} of $\vshapet$ with respect to a graph $G$. If $\vshapet$ is empty, no node in any graph is in the target of $\vshapet$. 
The definition of whether a node conforms to a shape, as we previously discussed, does not only depend on the graph $G$, but it might also depend, due to shape references, on whether other nodes conform to other shapes. 
Intuitively, the concept of \emph{assignments} \cite{Corman2018SHACL} is used to keep track, for every \trdf node, of all the shapes that it conforms to, and all of those that it does not. 
Given a document \vShapeDocument\ and a graph $G$, we denote $\nodesof{G}{\vShapeDocument}$ the set of nodes in $G$ together with any extra ones referenced by the node target declarations in \vShapeDocument . With \shapesof{\vShapeDocument} we refer to all the shape names in a document \vShapeDocument .

\begin{definition}
Given a graph $G$, and a \tshacl document \vShapeDocument , an \emph{assignment} $\sigma$ for $G$ and $\vShapeDocument$ is a function mapping nodes in \nodesof{G}{\vShapeDocument}, to subsets of $\shapesof{\vShapeDocument} \cup \{\neg \vshape | \vshape \in \shapesof{\vShapeDocument}\}$, such that for all nodes \vn\ and shape names \vshape ,  $\sigma(\vn)$ does not contain both \vshape\ and $\neg \vshape$.
\end{definition}

Expression $\GsigmaModels{G}{\sigma}{\vshapec}{\vn}$ denotes the evaluation of constraint $\vshapec$ on a node $\vn$ w.r.t.\ a graph $G$ under an assignment $\sigma$, as defined in \cite{Corman2018SHACL}.  If $\GsigmaModels{G}{\sigma}{\vshapec}{\vn}$ is \vtrue\ (resp.\ \vfalse) we say that node $n$ satisfies (resp.\ does not satisfy) constraint $\vshapec$ w.r.t.\ $G$ under $\sigma$. For any graph $G$ and assignment $\sigma$, fact $\vshape \in \sigma(\vn)$ (resp.\ $\neg \vshape \in \sigma(\vn)$) denotes the fact that node $\vn$ conforms (resp.\ does not conform) to $\vshape$ w.r.t.\ $G$ under $\sigma$. 
Expression $\GsigmaModels{G}{\sigma}{\vshapec}{\vn}$ evaluates to \vtrue , \vfalse\ or \vundefined\ values of Kleene's 3-valued logic, and the truth value of any shape reference in $\vshapec$ is computed using the assignment (it should be noted that the \vundefined\ value never occurs in non-recursive shapes, but it is used to define possible extended semantics in the face of recursion). In other words, whenever a truth value in the evaluation of $\GsigmaModels{G}{\sigma}{\vshapec}{\vn}$ depends on whether another node $\vj$ conforms to a shape $\vshape'$, with constraints $\vshapec'$, this is not resolved by evaluating $\GsigmaModels{G}{\sigma}{\vshapec'}{\vj}$, but instead it is  \vtrue\ if $\vshape' \in \sigma(\vj)$, \vfalse\ if $\neg \vshape' \in \sigma(\vj)$, or else \vundefined . This, in turn, eliminates the problem of a potentially infinite series of constraint evaluations.

The semantics of \tshacl validation can be defined with respect to a particular type of assignments, called \emph{faithful} \cite{Corman2018SHACL}.

\begin{definition} \label{def:faithfulAssignment}
For all graphs $G$, \tshacl documents $\vShapeDocument$ and assignments $\sigma$, assignment $\sigma$ is \emph{faithful} w.r.t. $G$ and $\vShapeDocument$, denoted with \isfaithful{G}{\sigma}{\vShapeDocument}, if the following two conditions hold for any shape $\trip{\vshape}{\vshapet}{\vshapec}$ in $\shapesof{\vShapeDocument}$ and node $\vn$ in \nodesof{G}{\vShapeDocument} :

(1) $\vshape \in \sigma(n)$ iff $\GsigmaModels{G}{\sigma}{\vshapec}{\vn}$ is \vtrue; and $\neg \vshape \in \sigma(n)$ iff $\GsigmaModels{G}{\sigma}{\vshapec}{\vn}$ is \vfalse;

(2) if $G \models \vshapet(\vn)$ then $\vshape \in \sigma(n)$.

\end{definition}

Condition (1) ensures that the facts denoted by the assignment are correct; while condition (2) ensures that the assignment is compatible with the target definitions. Condition (2) is trivially satisfied for \tshacl documents where all target definitions are empty.
Later we will want to discuss assignments where the first property of Def. \ref{def:faithfulAssignment} holds, but not necessarily the second, in order to reason about the existence of alternative assignments that are correct (as in, they satisfy the first part of Def. \ref{def:faithfulAssignment}) but that are not faithful. In fact, these will be faithful assignments to a document that is ``stripped empty'' of target definitions. Let $\minustargets{\vShapeDocument}$ denote the \tshacl document obtained from substituting all target definitions in \tshacl document $\vShapeDocument$ with the empty set. The following lemma holds:

\begin{lemma} \label{lemma:supportedModel}
For all graphs $G$, \tshacl documents $\vShapeDocument$ and assignments $\sigma$, condition (1) from Definition \ref{def:faithfulAssignment} holds for any shape $\vshape$ in $\shapesof{\vShapeDocument}$ and node $\vn$ in \nodesof{G}{\vShapeDocument} iff $\isfaithful{G}{\sigma}{\minustargets{\vShapeDocument}}$. 
\end{lemma}

The existence of a faithful assignment is a necessary and sufficient condition for validation for non-recursive \tshacl documents \cite{Corman2018SHACL}. As we will see later, this is also necessary condition for all the other extended semantics.
\begin{definition} \label{validationDefNonRecursive}
A graph $G$ is valid w.r.t.\ a non-recursive \tshacl document $\vShapeDocument$ if there exists an assignment $\sigma$ such that $ \isfaithful{G}{\sigma}{\vShapeDocument}$.
\end{definition}

\section{\tshacl Recursion}\label{sec:recursion}

The semantics of recursion in \tshacl documents is left undefined in the \tshacl specification \cite{2017SHACL}, and this gives rise to several possible interpretations. In this section we consider \emph{extended} semantics of \tshacl that define how to validate graphs against recursive \tshacl documents. We focus on existing extended semantics that follow monotone reasoning. These can be characterised by two dimensions, namely the choice between \emph{partial} and \emph{total} assignments \cite{Corman2018SHACL} and between \emph{brave} and \emph{cautious} validation \cite{SHACLstableModelSemantics}, which we will subsequently define. Put together, these two dimensions define the four extended semantics of \emph{brave-partial}, \emph{brave-total}, \emph{cautious-partial} and \emph{cautious-total}.
We will not go into the details of the less obvious dimension of \emph{stable-model} semantics \cite{SHACLstableModelSemantics}, which relates \tshacl to non-monotone reasoning in logic programs. 

As mentioned in the previous section, assignments can specify a truth value of \vtrue , \vfalse\ or \vundefined\ to whether a node conforms to given shape. The truth value of \vundefined , which does not occur in non-recursive \tshacl documents, can instead play an important role in validating \tshacl under recursion. Intuitively, this happens during validation, when recursion makes it impossible for a node $\vn$ to either conform or not to conform to a shape $\vshape$ but, at the same time, validity does not depend on whether $\vn$ conforms to shape $\vshape$ or not. Consider for example the following \tshacl document, containing a single shape \trip{\vshape^{*}}{\emptyset}{\vshapec^{*}} (with name \uri{InconsistentS} in this example). 
This shape is defined as the negation of itself, that is, given a node $\vn$, a graph $G$ and an assignment $\sigma$, fact $\GsigmaModels{G}{\sigma}{\vshapec^{*}}{\vn}$ is true iff $\neg \vshape^{*} \in \sigma(\vn)$, and false iff $ \vshape^{*} \in \sigma(\vn)$.

\begin{lstlisting}
:InconsistentS a sh:NodeShape ;
   sh:not :InconsistentS .
\end{lstlisting}

It is easy to see that any assignment that maps a node to either $\vshape^{*}$ or $\neg \vshape^{*}$ is not faithful, as it would violate condition (1) of Definition \ref{def:faithfulAssignment}. However, an assignment that maps every node of a graph to the empty set would be faithful for that graph and document $\{\vshape^{*}\}$. Intuitively, this means that nodes in the graph cannot conform nor not conform to shape $\vshape^{*}$, but since this shape does not have any target node to validate, then the graph can still be valid. The fact of whether nodes conform or not conform to shape $\vshape^{*}$ can thus be left as ``undefined''. 

This type of validation, for recursive \tshacl documents, is called validation with partial assignments. More specifically, validation under brave-partial semantics simply extends the criterion of Def. \ref{validationDefNonRecursive} to recursive \tshacl documents. All other extended semantics are constructed by adding additional conditions to brave-partial semantics.  The term ``partial'' should not be interpreted as the fact that it describes only ``part'' of nodes of a graph, or that it describes the relationship of a node to only ``part'' of the shapes. Within a partial assignment, the conformance of every node to every shape is precisely specified by one of three truth values, and the term ``partial'' only indicates that one of these three truth values is \vundefined.
\begin{definition} \label{validationDefPartialBrave}
A graph $G$ is valid w.r.t.\ a \tshacl document $\vShapeDocument$ under brave-partial semantics if there exists an assignment $\sigma$ such that $ \isfaithful{G}{\sigma}{\vShapeDocument}$.
\end{definition}

In the \tshacl specification, nodes either conform to, or not conform to a given shape, and the concept of an ``undefined'' level of conformance is arguably alien to the specification. It is natural, therefore, to consider restricting the evaluation of a constraint to the \vtrue\ and \vfalse\ values of boolean logic. This is achieved by restricting assignments to be \emph{total}. 
\begin{definition}
An assignment $\sigma$ is \emph{total} w.r.t. a graph $G$ and a \tshacl document $\vShapeDocument$ if, for all nodes $n$ in $\nodesof{G}{\vShapeDocument}$ and shapes $\trip{\vshape}{\vshapet}{\vshapec}$ in $\vShapeDocument$, either $\vshape \in \sigma(n)$ or $\neg \vshape \in \sigma(n)$.
\end{definition}

For any graph $G$ and \tshacl document $\vShapeDocument$ we denote with $\assignmentsPartial{G}{\vShapeDocument}$ and $\assignmentsTotal{G}{\vShapeDocument}$, respectively, the set of assignments, and the set of total assignments for $G$ and $\vShapeDocument$. Trivially, $\assignmentsTotal{G}{\vShapeDocument} \subseteq \assignmentsPartial{G}{\vShapeDocument}$ holds.

\begin{definition} \label{validationDefTotalBrave}
A graph $G$ is valid w.r.t.\ a \tshacl document $\vShapeDocument$ under brave-total semantics if there exists an assignment $\sigma$ in $\assignmentsTotal{G}{\vShapeDocument}$ such that $ \isfaithful{G}{\sigma}{\vShapeDocument}$.
\end{definition}

Since total assignments are a more specific type of assignments, if a graph $G$ is valid w.r.t.\ a \tshacl document $\vShapeDocument$ under brave-total semantics, than it is also valid w.r.t.\ $\vShapeDocument$ under brave-partial semantics. The reverse, instead, is only true for non-recursive \tshacl documents. In fact, as shown in \cite{Corman2018SHACL}, if there exists a faithful assignment for a graph $G$ and a non-recursive document $\vShapeDocument$, then there exists also a total faithful assignment for $G$ and $\vShapeDocument$. Therefore, the definition of validity under brave-total semantics (Def. \ref{validationDefTotalBrave}), for non-recursive \tshacl documents, coincides with the standard definition of validation (Def. \ref{validationDefNonRecursive}). 

While total assignments can be seen as a more natural way of interpreting the \tshacl specification, they are not without issues when recursive \tshacl documents are considered. Going back to our previous example, we can notice that there cannot exist a total faithful assignment for the \tshacl document containing shape \uri{InconsistentS}, for any non-empty graph. This is a trivial consequence of the fact that no node can conform to, nor not conform to, shape \uri{InconsistentS}. 
This, however, is in contradiction with the \tshacl specification, which implies that a \tshacl document without target declarations in any of its shapes (such as the one in our example) should trivially validate any graph. If there are no target declarations, in fact, there are no target nodes on which to verify the conformance of certain shapes, and thus no violations should be detected. 

The second and last dimension that we consider is the difference between brave and cautious validation. When a \tshacl document $\vShapeDocument$ is recursive, there might exist multiple assignments $\sigma$ satisfying property (1) of definition \ref{def:faithfulAssignment}, that is, such that $\isfaithful{G}{\sigma}{\minustargets{\vShapeDocument}}$. Intuitively, these can be seen as equally ``correct'' assignments with respect to the constraints of the shapes, and brave validation only checks whether at least one of them is compatible with the target definitions of the shapes. Cautious validation, instead, represents a stronger form of validation, where all such assignments must be compatible with the target definitions. 
\begin{definition} \label{ref:cautiousValidation}
A graph $G$ is valid w.r.t. a \tshacl document $\vShapeDocument$ under \emph{cautious-partial} (resp.\ \emph{cautious-total}) semantics if it is (1) valid under \emph{brave-partial} (resp.\ \emph{brave-total}) semantics and (2) for all assignments $\sigma$ in $\assignmentsPartial{G}{\vShapeDocument}$ (resp.\ $\assignmentsTotal{G}{\vShapeDocument}$), it is true that if $\isfaithful{G}{\sigma}{\minustargets{\vShapeDocument}}$ holds then $ \isfaithful{G}{\sigma}{\vShapeDocument}$ also holds.
\end{definition}

To exemplify this distinction, consider the following \tshacl document $\vShapeDocument_1$. This document requires the daily special of a restaurant, node \uri{DailySpecial}, to be vegetarian, that is, to conform to shape \uri{VegDishShape}. This shape is recursively defined as follows. Something is a vegetarian dish if it contains an ingredient, and all of its ingredients are vegetarian, that is, entities conforming to the \uri{VegIngredientShape}. A vegetarian ingredient, in turn, is an ingredient of at least one vegetarian dish.
\begin{lstlisting}
:VegDishShape a sh:PropertyShape ;
   sh:targetNode :DailySpecial ;
   sh:path :hasIngredient ;
   sh:minCount 1 ;
   sh:qualifiedMaxCount 0 ;
   sh:qualifiedValueShape [ sh: not :VegIngredientShape ] .

:VegIngredientShape a sh:PropertyShape ;
   sh:path [ sh:inversePath :hasIngredient ] ;
   sh:node :VegDishShape .
\end{lstlisting}
Consider now a graph $G_1$ containing the following triple.
\begin{lstlisting}
:DailySpecial :hasIngredient :Chicken .
\end{lstlisting}
Due to the recursive definition of \uri{VegDishShape}, there exist two different assignments $\sigma_1$ and $\sigma_2$, which are both faithful for $G_1$ and $\minustargets{\vShapeDocument_1}$. In $\sigma_1$, no node in $G_1$ conforms to any shape, while $\sigma_2$ differs from $\sigma_1$ in that node \uri{DailySpecial} conforms to \uri{VegDishShape} and node \uri{Chicken} conforms to \uri{VegIngredientShape}. 
Essentially, either both the dish and the ingredient from graph $G_1$ are vegetarian, or neither is. Therefore, $\sigma_2$ is faithful for $G_1$ and $\vShapeDocument_1$, while $\sigma_1$ is not. The question of whether the daily special is a vegetarian dish or not can be approached with different levels of ``caution''. Under brave validation, graph $G_1$ is valid w.r.t.\ $\vShapeDocument_1$, since it is possible that the daily special is vegetarian. Cautious validation, instead, takes the more conservative approach, and under its definition $G_1$ is not valid w.r.t.\ $\vShapeDocument_1$, since it is also possible that the daily special is not vegetarian. When analysing such recursive definitions, one might want to exclude ``unfounded'' assignments, that is, assignments that assign certain shapes to a node for no other reason than to allow the validation of a graph. This is achieved by the recursive semantics for \tshacl proposed in \cite{SHACLstableModelSemantics}, which is based on the concept of \emph{stable models} from Answer Set Programming. 

For each extended semantics, the definition of validity of a graph $G$ with respect to a \tshacl document $\vShapeDocument$, denoted by $G \models \vShapeDocument$, is summarised in the following list.
\begin{description}
    \item[\emph{brave-partial}] there exists an assignment that is faithful w.r.t. $G$ and $\vShapeDocument$;
    \item[\emph{brave-total}] there exists an assignment that is total and faithful w.r.t. $G$ and $\vShapeDocument$;
    \item[\emph{cautious-partial}] there exists an assignment that is faithful w.r.t. $G$ and $\vShapeDocument$, and every assignment that is faithful w.r.t. $G$ and $\minustargets{\vShapeDocument}$ is also faithful w.r.t. $G$ and $\vShapeDocument$.
    \item[\emph{cautious-total}] there exists an assignment that is total and faithful w.r.t. $G$ and $\vShapeDocument$, and every assignment that is total and faithful w.r.t. $G$ and $\minustargets{\vShapeDocument}$ is also faithful w.r.t. $G$ and $\vShapeDocument$.
\end{description}

\section{Formal languages for \tshacl}\label{sec:formalLanguages}

In this section we review the two main formal languages that have been proposed to model the semantics of \tshacl. We first discuss a complete first-order formalisation of \tshacl , which can be used to study a number of decision problems. We then present a simplified language that effectively models \tshacl constraints for the purpose of validation.

\subsection{\SCL , A First-Order Language for \tshacl}\label{sec:SCL}

In order to formally study \tshacl, it is convenient to abstract away from the syntax of its \trdf and \tsparql representations. The \SCL\ first order language \cite{pareti2020,pareti2021satisfiability} is currently the only complete formalisation of \tshacl into a formal logical system. The expressiveness of this language covers all of the \tshacl target declarations and all of the \tshacl core constraint components, including the filter components, which are less commonly studied. This language captures the semantics of whole \tshacl documents, and it can be used to study a number of related decision problems, including validation. 
The relation between \tshacl and \SCL\ is given by translation $\tau$ \cite{pareti2021satisfiability}, such that, given a \tshacl document $\vShapeDocument$, the first order sentence $\tau(\vShapeDocument)$ is the translation of $\vShapeDocument$ into \SCL . We identify the inverse translation with $\tau^{-}$.

Before defining \SCL\ and its properties, we must define how \trdf graphs and assignments are modelled in this logical framework. The domain of discourse is assumed to be the set of \trdf terms. Triples are modelled as binary relations, with atom $R(s,o)$ corresponding to triple $\tripsquare{s}{R}{o}$. A minus sign identifies the \emph{inverse} role, i.e.\ $R^{-}(s,o) = R(o,s)$. Binary relation name \isA represents class membership triples $\tripsquare{s}{\texttt{rdf:type}}{o}$ as $\isA(s,o)$. 
Assignments are modelled with a set of monadic relations \hasshapePredicate , called \emph{shape relations}. Each \tshacl shape $\vshape$ is associated with a unique shape relation $\hasshapePredicateS{\vshape}$ in \SCL. Facts $\hasshapePredicate(\vx)$ (resp. $\neg \hasshapePredicate(\vx)$) describe an assignment $\sigma$ such that $\vshape \in \sigma(\vx)$ (resp. $ \neg \vshape \in \sigma(\vx)$). Since this logical framework adopts boolean logic, $\forall \vx . \; \hasshapePredicate(\vx) \vee \neg \hasshapePredicate(\vx)$ holds, by the law of excluded middle. Thus shape relations define total assignments.

Given a graph $G$ and an assignment $\sigma$, we now define their respective translations \induced{$G$} and \induced{$\sigma$} into first order structures.

\begin{definition}
Given a graph $G$, fact $p(s,o)$ is true in the first order structure \induced{$G$} iff $\tripsquare{s}{p}{o} \in G$.
\end{definition}
\begin{definition}
Given a total assignment $\sigma$, fact $\hasshape{\vn}{\sconst}$ is true in the first order structure \induced{$\sigma$} iff $\sconst \in \sigma(\vn)$.
\end{definition}
\begin{definition}
Given a graph $G$ and a total assignment $\sigma$, the first order structure $I$ \emph{induced} by $G$ and $\sigma$ is the disjoint union of structures \induced{$G$} and \induced{$\sigma$}. Given a first order structure $I$: (1) the graph $G$  \emph{induced} by $I$ is the graph that contains triple \tripsquare{s}{p}{o} iff $I \models p(s,o)$ and (2) the assignment $\sigma$ induced by $I$ is the assignment such that, for all nodes $\vn$ and shape relations $\hasshapePredicateS{\vshape}$, fact $\vshape \in \sigma(\vn)$ is true iff $I \models \hasshape{\vn}{\sconst}$ and $\neg \vshape \in \sigma(\vn)$ iff $I \not\models \hasshape{\vn}{\sconst}$.
\end{definition}

The existence of faithful assignments using \SCL\ and its standard model-theoretic semantics is presented in the following theorem \cite{pareti2020}. Trivially, this also defines what condition, in \SCL, corresponds to validation under the brave-total extended semantics (Def. \ref{validationDefTotalBrave}), which also defines validation for non-recursive \tshacl documents (Def. \ref{validationDefNonRecursive}).

\begin{theorem} \label{centralSCL2SHACLCorrespondenceTheorem}
For any graph $G$, total assignment $\sigma$ and \tshacl document $\vShapeDocument$, it is true that $\isfaithful{G}{\sigma}{\vShapeDocument}$ iff $I \models \tau(\vShapeDocument)$, where $I$ is the first order structure induced by $G$ and $\sigma$. 

For any first order structure $I$ and \SCL\ formula $\phi$, it is true $I \models \phi$ iff $\isfaithful{G}{\sigma}{\tau^{-}(\phi)}$, where $G$ and $\sigma$ are, respectively, the graph and assignment induced by $I$.
\end{theorem}

Sentences in the \SCL\ language follow the $\varphi$ grammar in Definition \ref{def:synGeneral}. 

\begin{definition}
    \label{def:synGeneral}
    The \emph{SHACL} first order language (\SCL, for short) is the set of first order \emph{sentences} built according to the following context-free grammar, where $\conc$ is a constant from the domain of \trdf terms, $\hasshapePredicate$ is a shape relation, $F$ is a filter relation, with shape relations disjoint from filter relations, 
    $R$ is a binary-relation name,  $^{\star}$ indicates the transitive closure of the relation induced by $\pi(\vx,\vy)$, the superscript $\pm$ refers to a relation or its inverse, and $n \in \mathbb{N}$.
    \begin{align*}
\varphi \gramdef\; & 
      \top
      \mid \varphi \wedge \varphi
      \\ & \mid \hasshapenosubscript{\conc}{\sconst} 
      \mid \forall \vx \,.\, \isA(\vx, \conc) \rightarrow \hasshapenosubscript{\vx}{\sconst} 
      \mid  \forall \vx, \vy \,.\, R^{\pm}(\vx, \vy) \rightarrow \hasshapenosubscript{\vx}{\sconst} 
      \\ & \mid \forall \vx . \; \hasshapenosubscript{\vx}{\sconst}  \leftrightarrow \psi(\vx) ; \\
\psi(\vx) \gramdef\;
        & \top 
        \mid \neg \psi(\vx)
        \mid \psi(\vx) \wedge \psi(\vx) 
        \mid \vx = \conc 
        \mid F(\vx)
        \mid \hasshapenosubscript{\vx}{\sconst}  
        \mid \exists \vy .\, \pi(\vx, \vy) \wedge \psi(\vy)
        \\ & \mid \neg \exists \vy .\, \pi(\vx, \vy) \wedge R(\vx, \vy)
          & {\normalfont \textsf{\textbf{[$\D$]}}}
        \\ & \mid \forall \vy .\, \pi(\vx, \vy)
          \leftrightarrow R(\vx, \vy)  & {\normalfont \textsf{\textbf{[$\E$]}}}
        \\ & \mid \forall \vy, \vz \,.\, \pi(\vx, \vy) \wedge R(\vx, \vz) \rightarrow \sigmagrammar(\vy, \vz) & {\normalfont \textsf{\textbf{[$\O$]}}}
        \\ & \mid \exists^{\geq n} \vy \,.\, \pi(\vx, \vy) \wedge \psi(\vy)  ; & {\normalfont \textsf{\textbf{[$\C$]}}} \\
\pi(\vx, \vy) \gramdef\;
        & R^{\pm}(\vx, \vy) 
        \\ & \mid \exists \vz \,.\, \pi(\vx, \vz) \!\wedge\!
          \pi(\vz, \vy) & {\normalfont \textsf{\textbf{[$\S$]}}} 
        \\ & \mid \vx \!=\! \vy \!\vee\! \pi(\vx, \vy) & {\normalfont \textsf{\textbf{[\Z]}}} 
        \\ & \mid \pi(\vx, \vy) \!\vee\! \pi(\vx, \vy) & {\normalfont \textsf{\textbf{[$\A$]}}} 
        \\ & \mid  (\pi(\vx, \vy))^{\star} ; & {\normalfont \textsf{\textbf{[\T]}}} \\ 
\sigmagrammar(\vx, \vy) \gramdef\;
        & \vx <^{\pm} \vy \mid \vx \leq^{\pm} \vy.  
    \end{align*}
\end{definition}

Symbol $\varphi$ corresponds to a \tshacl document. An \SCL\ sentence could be empty ($\top$), a conjunction of documents, a \emph{target axiom} representing a target definition (a production of the 3rd, 4th and 5th production rule) or a \emph{constraint axiom} representing a constraint (a production of the last production rule). Target axioms take one of three forms, based on the type of target declarations. The translation of \tshacl target declarations into \SCL\ target axioms is summarised in Table~\ref{tab:translationTargs}.
Letters in square brackets are annotations for naming \SCL\ components and thus are not part of the grammar. These letters are essentially first-letter abbreviations of \emph{prominent} \tshacl components (that together define fragments of \SCL), and are also listed in Table \ref{tab:SHACLcomponentsInOurGrammar}. 

\begin{table}[t]
\begin{center}
\caption{ 
Translation of a shape with name $\vshape$ with a target definition $\vshapet$, into an \SCL\ target axiom.
} 
\label{tab:translationTargs}
\begin{tabular}{ |l | l |}
 \hline
 Target declaration in $\vshapet$  & \SCL\ target axiom \\ \hline 
 Node target (node \conc) & $\hasshape{\conc}{\vshape}$ \\ \hline
 Class target (class \texttt{c})  & $\forall \vx . \isA(\vx,$\texttt{c}$)\rightarrow \hasshape{\vx}{\vshape}$\\ \hline
 Subjects-of target (relation $R$)  & $\forall \vx, \vy . R(\vx,\vy) \rightarrow \hasshape{\vx}{\vshape}$\\ \hline
 Objects-of target (relation $R$)  & $\forall \vx, \vy . R^{-}(\vx,\vy) \rightarrow \hasshape{\vx}{\vshape}$ \\ \hline
\end{tabular}
\end{center}
\end{table}

The non terminal symbol $\psi(\vx)$ corresponds to the subgrammar of the \tshacl constraints components. 
Within this subgrammar, $\top$ identifies an empty constraint, $\vx = \conc$ a constant equivalence constraint and $F$ a monadic filter relation (e.g.\ $F^{\isIRI}(\vx)$, true iff $\vx$ is an IRI). 
Filters components are captured by $F(\vx)$ and the \lO\ component. The \C\ component captures qualified value shape cardinality constraints. The \E, \D\ and \lO\ components capture the equality, disjointedness  and order property pair components. 
The $\pi(\vx, \vy)$ subgrammar models \tshacl property paths. Within this subgrammar \S\ denotes sequence paths, \A\ denotes alternate paths, \Z\ denotes a zero-or-one path and \T\ denotes a zero-or-more path.

Translation $\tau$ results in a subset of \SCL\ formulas, called \emph{well-formed} defined subsequently, and the inverse translation $\tau^{-}$ only takes well formed sentences as an input. An \SCL\ formula $\phi$ is well-formed iff for every shape relation $\Sigma$, formula $\phi$ contains exactly one constraint axiom with relation $\Sigma$ on the left-hand side of the implication. Intuitively, this condition ensures that every shape relation is ``defined'' by a corresponding constraint axiom. The translation of the document from Fig. \ref{fig:exampleSHACL2}, into a well-formed \SCL\ sentence, via $\tau$, is the following. Arguably, this logic notation might seem easier to read and understand than the SHACL syntax of Fig. \ref{fig:exampleSHACL2}. 

\begin{align*}
&\Big( \forall \vx . \;  \isA(\vx, \texttt{:Employee}) \rightarrow \hasshape{\vx}{\texttt{:EmployeeShapeB}} \Big) \\
&\quad \wedge \Big( \forall \vx . \; \hasshape{\vx}{\texttt{:EmployeeShapeB}} \leftrightarrow   \exists \vy . \; R_{\texttt{:hasOfficeNumber}}(\vx, \vy) \wedge \hasshape{\vy}{\texttt{:OfficeNumberShape}} \Big) \\
&\quad \wedge \Big( \forall \vx . \; \hasshape{\vx}{\texttt{:OfficeNumberShape}} \leftrightarrow   F^{\minLength 3}(\vy) \Big) 
\end{align*}

\begin{table*}[t]
\begin{center}
\caption{Relation between prominent \tshacl components and \SCL\ expressions.}\label{tab:SHACLcomponentsInOurGrammar}
\begin{tabular}{ | l | l |l | l |}
 \hline
 Abbr. & Name  & \tshacl component & Corresponding expression  \\ \hline
 \S & Sequence Paths & Sequence Paths & $\exists \vz \,.\, \pi(\vx,\vz) \wedge \pi(\vz,\vy)$ \\ \hline
 \Z & Zero-or-one  Paths  & \sh{zeroOrOnePath} & $\vx = \vy \vee \pi(\vx, \vy)$ \\ \hline
  \A & Alternative Paths & \sh{alternativePath} & $\pi(\vx, \vy) \vee \pi(\vx, \vy)$ \\ \hline
  \T & Transitive Paths  & \begin{tabular}{@{}l@{}}\sh{zeroOrMorePath} \\ \sh{oneOrMorePath}
 \end{tabular}  & $(\pi(\vx, \vy))^{\star}$ \\ \hline
 \D & Property Pair Disjointness & \sh{disjoint} & $\neg \exists \vy . \pi(\vx, \vy) \wedge R(\vx, \vy)$ \\ \hline
 \E & Property Pair Equality & \sh{equals} & $\forall \vy \,.\, \pi(\vx,\vy) \leftrightarrow R(\vx,\vy)$ \\ \hline
 \O & Property Pair Order  & \begin{tabular}{@{}l@{}}\sh{lessThan} \\\sh{lessThanOrEquals}\end{tabular} & \begin{tabular}{@{}l@{}}$\vx \leq^{\pm} \vy \text{ and } \vx <^{\pm} \vy $ 
 \end{tabular} \\ \hline
 \C & Cardinality Constraints  & \begin{tabular}{@{}l@{}}\sh{qualifiedValueShape} \\ \sh{qualifiedMinCount} \\ \sh{qualifiedMaxCount}\end{tabular}   & \begin{tabular}{@{}l@{}}$\exists^{\geq n} \vy \,.\, \pi(\vx, \vy) \wedge \psi(\vy)$ \\ with $n \not = 1$\end{tabular}\\ \hline
\end{tabular} 
\end{center}
\end{table*} 

The language defined without any of these constructs is called the \emph{base} language, denoted \X. On top of the base language different syntactic fragments of \SCL\ are defined by considering different combinations of features allowed. We name these fragments by concatenating the letters that represent the features allowed, into a single name. For example, \S \A\ identifies the fragment that only allows the base language, sequence paths and alternate paths. This means that in order to write an \SCL\ document in \S \A,  
one can only use the production rules of Def.~\ref{def:synGeneral} that are not annotated with any feature (base language) or those identified by abbreviations \S\ and \A.
 
The \tshacl specification presents an unusual asymmetry in the fact that equality, disjointedness and order components force one of their two path expressions to be an atomic relation. 
This can result in situations where the order constraints can be defined in just one direction, since only the less-than and less-than-or-equal property pair constraints are defined in \tshacl. 
The \O\ fragment models a more natural order comparison that includes the $>$ and $\geq$ components. The fragment where the order relations in the $\sigmagrammar(\vx, \vy)$ subgrammar cannot be inverted is denoted \O'.
 
When interpreting an \SCL\ sentence, particular care should be paid to the semantics of filter relation. The interpretation of each filter relation, such as $F^{\isIRI}(\vx)$, is the subset of the domain of discourse on which the filter is true. This interpretation is constant across all models, and defines the semantics of the filter. When considering the decision problem of validation, filter relations in \SCL\ must be suitably defined by interpreted relations (similarly to how the equality operator is). When considering additional decision problems, such as satisfiability and containment (which will be discussed in Section \ref{sec:SHACLdecisionProblems}), the semantics of filters can be axiomatisatised, thus removing the need for special interpreted relations. The filter axiomatisation presented in \cite{pareti2020} captures the semantics of all SHACL filters with the single exception of \sh{pattern}, as this filter defines complex non-standard regular expressions based on the \tsparql \tregex function \cite{SeaborneSPARQL2013}.

\subsection{$\mathcal{L}$, a Language for \tshacl Constraint Validation }

Another major language used to study \tshacl is $\mathcal{L}$ which was presented in \cite{Corman2018SHACL} and paved the way to subsequent formal studies of \tshacl. The $\mathcal{L}$ language differs from \SCL\ in scope and purpose. While \SCL\ sentences describe whole \tshacl documents, sentences in $\mathcal{L}$ describe individual \tshacl constraints. The $\mathcal{L}$ language is primarily designed to investigate the complexity of \tshacl validation. As such, it relies on assumptions that do not hold when studying other decision problems such as satisfiability and containment, which, instead, can be studied using \SCL . In particular, $\mathcal{L}$ assumes that all filter components can be evaluated on a node in constant time, and thus are all equivalent, for the purposes of validation. Thanks to this reduced scope, $\mathcal{L}$ seems less complex than \SCL , and it is a useful formalism to study the evaluation of \tshacl constraints. The semantics of an $\mathcal{L}$ sentence $\phi$ is defined in \cite{Corman2018SHACL} through the use of faithful assignments. In particular, \cite{Corman2018SHACL} fixes a lookup table that provides the truth value of the evaluation of $\phi$ on a node $\vn$ for a graph $G$ and an assignment $\sigma$. Instead, \SCL\ relies on the standard model-theoretic semantics. 

\begin{table*}[t]
\begin{center}
\caption{Correspondence between an $\mathcal{L}$ sentence $\phi$, and \SCL\ $\psi^{\phi}(\vx)$ expressions, such that a constraint $\phi$ is satisfied on a node $\vn$ w.r.t. a graph $G$ and an assignment $\sigma$ iff $I \models \psi^{\phi}(\vn)$, where $I$ is the first order structure induced by $G$ and $\sigma$. It is assumed that paths are expressed using the $\pi(\vx,\vy)$ subgrammar of $\SCL$, and that $r_{2}$ is an IRI.}\label{tab:LtoSCLcorrespondence}
\begin{tabular}{ | l | l  |}
 \hline
 $\mathcal{L}$ expression $\phi$ & Corresponding \SCL\ $\psi^{\phi}(\vx)$ \\ \hline
 $\top$ & $\top$ \\ \hline
 $s$ & $\hasshape{\vx}{\sconst} $ \\ \hline
 $I$ & $\vx = I $ \\ \hline
 $\phi_{1} \wedge \phi_{2}$ & $\psi^{\phi_{1}}(\vx) \wedge \psi^{\phi_{2}}(\vx) $ \\ \hline
 $\neg \phi$ & $\neg \psi^{\phi}(\vx)$ \\ \hline
 $\geq_{n} r . \phi$ & $\exists^{\geq n} \vy \,.\, r(\vx, \vy) \wedge \psi^{\phi}(\vy)$ \\ \hline
 $\mathrm{EQ}\left(r_{1}, r_{2}\right)$ & $\forall \vy .\, r_{1}(\vx, \vy)
          \leftrightarrow r_{2}(\vx, \vy)$ \\ \hline
\end{tabular} 
\end{center}
\end{table*} 

The grammar of $\mathcal{L}$ sentences is given next. In this grammar $s$ is a shape name; $I$ is an IRI; $r$ is a \tshacl property path; $n$ is a positive integer. 
$$
\phi :=\top \, | \, s \, | \,  I  \,  | \, \phi_{1} \wedge \phi_{2}  \, | \,  \neg \phi  \, |  \, \geq_{n} r . \phi \, | \,  \mathrm{EQ}\left(r_{1}, r_{2}\right)
$$

Table \ref{tab:LtoSCLcorrespondence} defines the correspondence between $\mathcal{L}$ and the $\psi(\vx)$ sub-grammar of \SCL .  It is easy to see that $\mathcal{L}$ sentences correspond to a subset of the $\psi(\vx)$ sub-grammar of \SCL , assuming that $r_2$ denotes a predicate path. This assumption is required as in $\mathcal{L}$ both arguments of $\mathrm{EQ}\left(r_{1}, r_{2}\right)$, which captures the \tshacl equality operator (\sh{equals}), are path expressions. This is a generalisation of \tshacl, since the \tshacl specification requires one of the two paths to be a simple predicate path, or in other words, an IRI. It should also be noted that $\mathcal{L}$ does not model property pair order components (denoted \O\ in \SCL ), and that the \sh{closed} component is modelled using path expression operators not supported by \tshacl paths. The \tshacl disjoint constraint component (denoted \D\ in \SCL ) is only implicitly included in $\mathcal{L}$ when considering recursion. It is possible, in fact, to represent a disjoint constraint component in $\mathcal{L}$ using two auxiliary recursive shapes \cite{Corman2018SHACL}.

\section{\tshacl Decision Problems}\label{sec:SHACLdecisionProblems}

Several existing pieces of work in the literature focus on \tshacl, and several related decision problems have been investigated. In Section \ref{sec:shaclvalidation} we review existing work on the core decision problem for \tshacl, namely validation. Unlike validation, which studies the relationship between a \tshacl document and an \trdf graph, the decision problems of \emph{satisfiability} and \emph{containment}, reviewed in Section \ref{sec:shaclsatandcontainment}, focus on intrinsic properties of SHACL documents and their components. 

\subsection{Validation} \label{sec:shaclvalidation}

Validation is a core decision problem for \tshacl, since the main application of this language is the validation of \trdf graphs. This decision problem is decidable for all of the semantics discussed in this article, including the four extended semantics. The complexity lower bounds for validation, however, depend on the fragment of \tshacl being considered. Table \ref{tab:validationComplexity} lists the data complexity of three fragments of \tshacl given in \cite{Corman2018SHACL,SHACLstableModelSemantics}. The three fragments are (1) \tshaclnonrec, the fragment of non-recursive \tshacl documents built using $\mathcal{L}$ constraints; (2) \tshaclplus, the fragment of \tshacl documents built using $\mathcal{L}$ constraints with a restricted use of negation, that is, substituting the $\neg \phi$ production rule of $\mathcal{L}$ into $\phi_{1} \vee \phi_{2}$; and (3) \tshaclrec, the fragment of \tshacl documents built using $\mathcal{L}$ constraints. The most expressive of these fragments,  \tshaclrec,  is NP-complete in data complexity.

\begin{table*}[t]
\begin{center}
\caption{Data complexity of \tshacl validation, results from \cite{SHACL2SPARQLtranslation}.}\label{tab:validationComplexity}
\begin{tabular}{ | l | l  |}
 \hline
 Fragment & Data complexity of validation \\ \hline
 \tshaclnonrec & NL-c \\ \hline
 \tshaclplus & PTIME-c \\ \hline
 \tshaclrec & NP-c \\ \hline 
\end{tabular} 
\end{center}
\end{table*} 

\subsection{Satisfiability and Containment} \label{sec:shaclsatandcontainment}

Satisfiability and containment are standard decision problems that have been investigated in the context of \tshacl. These two decision problems, unlike validation, do not take a graph as an input. Instead, they focus on \tshacl documents, shapes or constraints. Given any notion of validity from one of the semantics defined earlier, the following decision problems are defined. For simplicity, when discussing satistiability and containment, we will assume the use of the semantics of validation from Definitions \ref{validationDefTotalBrave} and \ref{validationDefNonRecursive}.

\begin{definition}
A \tshacl document $\vShapeDocument$ is \emph{satisfiable} iff there exists a graph $G$ such that $G \models \vShapeDocument$. Deciding whether a \tshacl document is satisfiable is the decision problem of \tshacl \emph{satisfiability}.
\end{definition}
\begin{definition}
\textbf{\tshacl Containment}: For all \tshacl documents $\vShapeDocument_1$, $\vShapeDocument_2$, we say that $\vShapeDocument_1$ is \emph{contained} in $\vShapeDocument_2$, denoted $\vShapeDocument_1 \subseteq \vShapeDocument_2$, iff for all graphs $G$, if $ G \models \vShapeDocument_1$ then $G \models \vShapeDocument_2$. Deciding whether a \tshacl document is contained in another is the decision problem of \tshacl \emph{containment}.
\end{definition}
Two \tshacl documents $\vShapeDocument_1$ and $\vShapeDocument_2$ that are contained in each other ($\vShapeDocument_1 \subseteq \vShapeDocument_2$ and $\vShapeDocument_2 \subseteq \vShapeDocument_1$) are \emph{semantically equivalent}. Two semantically equivalent documents are not necessarily equivalent syntactically, since in \tshacl the same constraint can be expressed using different sets of shapes.

The satisfiability and containment decision problems for \tshacl can be polynomially reduced to the satisfiability decision problem for \SCL , defined as follows in the natural way \cite{pareti2020}. 
\begin{definition}
An \SCL\ sentence $\phi$ is satisfiable iff there exists structure $\Omega$ such that $\Omega \models \phi$.  Deciding whether a \SCL\ sentence is satisfiable is the decision problem of \SCL\ \emph{satisfiability}. 
\end{definition}

\begin{figure*}[t]
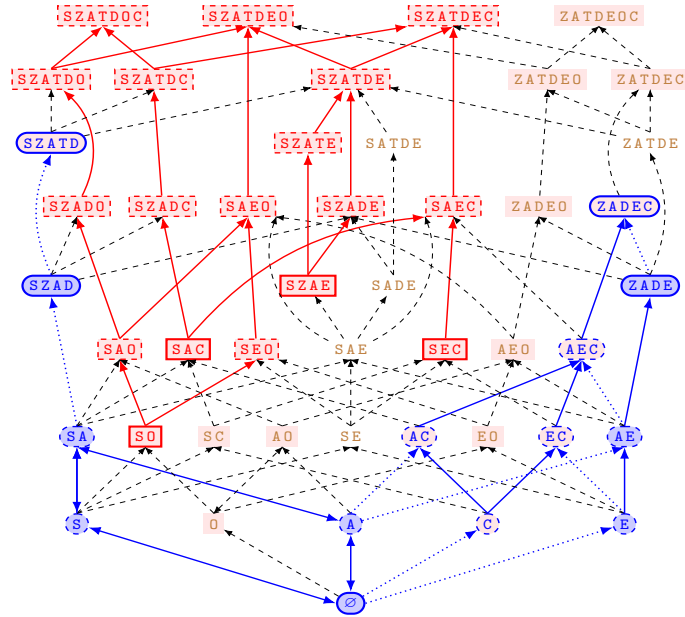

    \figfrg
    \caption{\label{fig:log} \cite{pareti2020} Decidability and complexity map of \SCL\ satisfiability.
      Round (blue) and square (red) nodes denote decidable and undecidable
      fragments, respectively.
      Solid borders on nodes correspond to theorems in this paper, while dashed
      borders are implied results.
      Directed edges indicate inclusion of fragments, while bidirectional edges
      denote polynomial-time reducibility.
      Solid edges are preferred derivations to obtain tight results, while
      dotted ones leads to worst upper-bounds or model-theoretic properties.
      Finally, a light blue background indicates that the fragment enjoys the
      finite-model property, while those with a light red background do not
      satisfy this property.} 
  \end{figure*}

This reduction allows us to study the decidability and complexity  of the \tshacl satisfiability and containment problems for a given \tshacl fragment by studying the decidability and complexity of \SCL\ satisfiability, for the corresponding fragments. The results of this study, published in \cite{pareti2020}, are summarised in Figure \ref{fig:log}. Negative results indicate the undecidability of both the \SCL\ fragment, and the corresponding \tshacl fragment. Positive results, shown in round blue in the figure, indicate that both satisfiability and containment are decidable, for that fragment of non-recursive \tshacl, and are accompanied with complexity upper-bounds. Starting from the negative results, \tshacl satisfiability and containment is, in general, undecidable. This was shown even for several non-recursive fragments, through a semi-conservative reduction from the standard domino problem~\cite{Wan61,Ber66,Rob71}, which is an undecidable decision problem. More specifically, the \tshacl satisfiability problems for the \S\,\lO, \S\,\A\,\C, \S\,\E\,\C,
    \S\,\E\,\lO', and \S\,\Z\,\A\,\E\ fragments are undecidable \cite{pareti2020}.

Positive results are obtained by noticing that several \SCL\ fragments are included in decidable fragments of first order logic. For example, the \S\,\Z\,\A\,\T\,\D\ fragment of \SCL\ is included in the extension of the unary-negation fragment of first-order logic with arbitrary transitive relations, which can be solved in 2ExpTime~\cite{ABBV16,JLMS18,DK19}. The complexity upper bounds identified in \cite{pareti2020} for \tshacl fragments range from ExpTime to 2ExpTime. 
Both decision problems are defined over \tshacl documents which, similarly to the schema of a dataset, could be assumed to be of small or constant size.

Up until this point we considered the satisfiability and containment problems defined at the level of \tshacl documents. However, it is possible to study variations of these problems at different levels of granularity. For example, the satisfiability and containment problems at the level of \tshacl constraints are defined in \cite{pareti2020}, and are shown to be reducible to the problem of \tshacl satisfiability. An approach that uses Description Logics Reasoning is presented in \cite{martin2020shapecontainment} to compute \emph{shape containment}, that is, containment at the level of shapes, for a restrictive fragment of \tshacl , which however allows recursion.

\section{Inference Rules and the Schema Expansion} \label{sec:schemaexpansion}

Datasets are often dynamic objects, which are frequently subject to modification. When an \trdf graph is modified, its validity w.r.t.\ a \tshacl document might change. If the modifications that a dataset undergoes are completely arbitrary, then it is not possible to make predictions regarding validity, and the dataset might need to be re-validated after each modification. Many types of modifications that can be applied on a dataset, however, are predictable or a result of some reasoning process. In particular, many types of modifications can be represented as \emph{inference rules} $B \rightarrow H$, where a set of facts $H$, called the \emph{head} are added to a dataset whenever a query $B$, called the \emph{body}, finds a match on the dataset. Given an \trdf graph $G$, and a set of inference rules $R$, it is possible to compute graph $G'$, \emph{closure} of $G$ under $R$, by applying the \emph{chase} algorithm \cite{benedikt2017benchmarking}. The chase algorithm, intuitively, consists in repeatedly applying the rules of $R$ on $G$ until convergence. For simplicity, we assume that the chase algorithm is guaranteed to terminate for the inference rules considered.

Assuming graph $G$ is valid w.r.t.\ to a \tshacl document $\vShapeDocument$, the approach presented in \cite{pareti2019c}, called \emph{schema expansion}, allows us to predict whether the graph closure $G'$ will still be valid w.r.t.\ $\vShapeDocument$ without having to validate $G'$ against $\vShapeDocument$. In particular, given a \tshacl document $\vShapeDocument$ and a set of inference rules $R$, the schema expansion process computes the ``maximal sub-document'' of $M$ which will still validate $G$ after the rule applications. That is, the schema expansion is a \tshacl document $\vShapeDocument'$, called \emph{schema consequence}, such that (1) $\vShapeDocument \subseteq \vShapeDocument'$ (i.e., $\vShapeDocument'$ is a subset of the restrictions of $\vShapeDocument$); (2) validity is preserved after closure, that is,
for any graph $G$ valid w.r.t.\ $\vShapeDocument$, its closure $G'$ under $R$ is valid w.r.t. $\vShapeDocument'$; and (3) $\vShapeDocument'$ is ``minimally-containing'', i.e., there is no document  $\vShapeDocument''$ that satisfies conditions (1) and (2) and such that that $\vShapeDocument'' \subset \vShapeDocument'$. 
If a schema consequence $\vShapeDocument'$ of a \tshacl document $\vShapeDocument$ under inference rules $R$ is semantically equivalent to $\vShapeDocument$, then any graph $G$, valid w.r.t.\ $\vShapeDocument$ is guaranteed to remain valid w.r.t.\ $\vShapeDocument$ after computing its closure under $R$. In other words, this means that the application of rules $R$ cannot ``invalidate'' graphs valid w.r.t.\ document  $\vShapeDocument$.

Consider, for example, the following graph $G_{1}$, which describes \uri{Eve}, a manager of the company in the IT department, and one of her subordinates \uri{Fiona}.
\begin{lstlisting}
:Eve a :Manager ;
    :hasDepartment "IT" .
:Fiona a :Employee ;
    :hasManager :Eve .
\end{lstlisting}
This graph is valid w.r.t.\ the following \tshacl document $\vShapeDocument_{1}$, which states that each employee must have a manager, and each manager must have a department.
\begin{lstlisting}
:SubordinateS a :PropertyShape ;
    sh:targetClass :Employee ;
    sh:path :hasManager  ;
    sh:minCount 1 .
    
:ManagerS a sh:PropertyShape ;
    sh:targetClass :Manager ;
    sh:path :hasDepartment  ;
    sh:minCount 1 .
\end{lstlisting}

Consider now the set of inference rules $R_{1} =\{r_1,r_2\}$, where rules $r_1$ and $r_2$ are defined as follows. For simplicity, we represent both the head and the body of rules as \tsparql graph patterns, which are interpreted as \tsparql \tconstruct queries where the \twhere and \tconstruct clauses are the body and the head, respectively. Rule $r_1$ states that every manager can be inferred to be an employee, and $r_2$ states that everyone can be inferred to be in the same department as their manager. 
\begin{lstlisting}[escapeinside={(*}{*)}]
(*$r_1 = $*) {?x rdf:type :Manager} (*$\rightarrow$*) {?x rdf:type :Employee}
(*$r_2 = $*) {?x :hasManager ?y . ?y :hasDepartment ?z} (*$\rightarrow$*) {?x :hasDepartment ?z}
\end{lstlisting}

The closure of graph $G_{1}$ under rules $R_{1}$ is the following graph $G_{2}$.
\begin{lstlisting}
:Eve a :Manager ;
    a :Employee ;
    :hasDepartment "IT" .
:Fiona a :Employee ;
    :hasManager :Eve ;
    :hasDepartment "IT" .
\end{lstlisting}
Notice that graph $G_{2}$ is not valid w.r.t.\ $\vShapeDocument_{1}$, since \uri{Eve} violates \uri{SubordinateS}, but it is valid w.r.t.\ another document $\vShapeDocument_{2}$ which only contains shape \uri{ManagerS}. In fact, $\vShapeDocument_{2}$ is a schema consequence of $\vShapeDocument_{1}$ and $R_{1}$. Therefore, we know that the closure under $R_{1}$ of any graph valid w.r.t.\ $\vShapeDocument_{1}$ will validate shape \uri{ManagerS}, but it might not validate \uri{SubordinateS}.

Two approaches to compute the schema expansion are presented in \cite{pareti2019c}, for datalog \cite{Ceri1989Datalog} inference rules without negation. The first based on the concept of \emph{critical instance} \cite{marnette2009generalized}, and the second an optimisation of the first. These approaches are only defined on a fragment of \tshacl that, although restricted, is sufficient to express common constraints for \trdf validation, such as the  Data Quality Test Patterns \textsc{typedep}, \textsc{typrodep}, \textsc{pvt}, \textsc{rdfs-domain} and \textsc{rdfs-range} in the categorisation by Kontokostas et al.\ 
\cite{Kontokostas2014RuleValidation}. Intuitively, the difficulty in computing a schema expansion lies in having to consider all possible graphs that are valid w.r.t.\ a \tshacl document, and their interactions with arbitrarily complex inference rules.

\section{Applications, Tools and Implementations} \label{sec:relatedwork}

Over a few years since reaching its status as a W3C recommendation, the level of maturity and adoption of the \tshacl technology has been steadily increasing. In this section we review existing \tshacl implementations, tools designed to facilitate the creation and management of \tshacl documents, and documented usages of \tshacl in practical applications.

\subsection{Tools for \tshacl Validation}

The availability of mature tools is often a crucial requirement for the widespread adoption of a technology. To date, \tshacl validation has been integrated in a number of mainstream tools and triplestores.\footnote{\url{https://w3c.github.io/data-shapes/data-shapes-test-suite/} accessed on \\ 18/6/21} 
An example of this is RDF4J,\footnote{\url{https://rdf4j.org/} accessed on 18/6/21} a Java framework for managing \trdf data, which now includes an engine for \tshacl validation. The RDF4J framework is integrated in a number of projects, most notably the GraphDB\footnote{\url{https://graphdb.ontotext.com/} accessed on 18/6/21} triplestore. Other \tshacl-enabled databases include AllegroGraph\footnote{\url{https://allegrograph.com/} accessed on 18/6/21} by Franz Inc, Apache Jena\footnote{\url{https://jena.apache.org/} accessed on 18/6/21} by Apache, and Stardog\footnote{\url{https://www.stardog.com/} accessed on 18/6/21} by Stardog Union Inc. A benchmark for the comparison of different \tshacl implementation was proposed in \cite{SHACLbenchmark2020}, along with results for four different databases. A \tshacl implementation is also available for Python through the pySHACL\footnote{\url{https://pypi.org/project/pyshacl/} accessed on 18/6/21} library.

One of the first tools to enable the validation of recursive \tshacl graphs was SHACL2SPARQL \cite{corman2019shacl2sparql}. Another tool, Trav-SHACL \cite{TravSHACL2021}, implements a \tshacl engine designed to optimise the evaluation of \tshacl core constraint components expressible in fragments of the $\mathcal{L}$ language \cite{Corman2018SHACL}. On these fragments of \tshacl, Trav-SHACL was shown to achieve significantly faster validation times compared to the SHACL2SPARQL tool.

\subsection{Tools for Generating \tshacl Documents}

While efficient tools to perform graph validation are undoubtedly essential to the widespread adoption of \tshacl, it is also important to devise practical ways to generate suitable \tshacl documents, without which validation would not be possible. On the one hand, \tshacl documents can be manually created by experts. Tools to support this manual process can facilitate this, especially when integrated with already established software. An example of this is SHACL4P \cite{7936162}, a plugin for the Protégé ontology editor \cite{Protege2015} which includes an editor to create \tshacl documents, and a validator that allows users to test the document by validating an ontology with it, and then visualising any constraint violations. Shape Designer \cite{bonevashapedesigner} is another tool to create \tshacl documents that combines a graphical editor, and additional algorithms to create constraints semi-automatically by analysing the data graph. The benefits of different types of visualisations as an aid to the creating and editing of constraints for \trdf graphs was studied in \cite{liebervisual}. Existing work also investigated the possibility of generating \tshacl documents from natural language text \cite{10.1007/978-3-030-35646-0_9}.
 
A number of approaches have been designed to automate the creation of \tshacl documents.  
The SHACLearner \cite{omranlearning} approach, generates \tshacl documents by learning a kind of rules called Inverse Open Path (IOP) Rules from the graph data provided. IOP rules are strongly related to \SCL\, and therefore \tshacl. An IOP rule essentially follows the same structure as an \SCL\ constraint axiom, both syntactically and semantically, with the only exception that the iff operator is replaced by a rightward implication. Another approach to automate the creation of \tshacl documents is the Astrea-KG Mappings \cite{astrea}. These mappings consist of a set of manually created mappings from OWL \cite{OWL2} to \tshacl, that can be used to automatically generate \tshacl documents from OWL ontologies. As described in \cite{Pandit2018UsingODP4SHACL}, \tshacl documents can also be generated from the axioms defined by ontology design patterns \cite{Gangemi2009}. 
The approach from \cite{spahiu2018towards} generates \tshacl documents for the purpose of quality assessment, using the ontology design patterns and data statistics created by the ABSTAT \cite{ABSTAT2016} tool as an input. Another similar approach, presented in \cite{ns2021schemabacked}, allows the automatic extraction of \tshacl constraints from a \tsparql endpoint, and was tested on the dataset of  Europeana\footnote{\url{https://www.europeana.eu/en} accessed on 18/6/21}.

Notably, \tshacl documents can also be seen as describing a desirable ``schema'' for graph data. As such, they can be used as a template to generate new RDF data. An example of this is the Sch{\'i}matos \cite{Schimatos2020} tool, which generates forms for RDF graph editing based on \tshacl documents, in order to simplify the graph editing task, and minimize the chance of error.

\subsection{Adoption of \tshacl}

An analysis of existing use cases of \tshacl can be useful to gain insights on how this technology is used in practice, and on in its level of adoption. 
In a recent review, 13 existing projects using \tshacl have been reviewed, and the most common constraints observed were cardinality, class, datatype and disjunction \cite{lieber2020statistics}. Several works investigate the use of \tshacl to verify compliance of a dataset w.r.t.\ certain policies, such as GDPR requirements \cite{shacl4gdpr,al2020automated}. Other applications of \tshacl include type checking program code \cite{MartinTypeCheckingCodeWithSHACL2019} and detecting metadata errors in clinical studies \cite{PMID:34042768}. \tshacl is also used by the European Commission to facilitate data sharing, for example by validating metadata about public services against the recommended vocabularies \cite{stani2019design}. Notably, several approaches define 
translations into \tshacl from other technologies, such as ontologies and other schema and constraint languages \cite{DiCiccio2019,k2019modellingConstruction,9205800,8249433,10.1007/978-3-030-21348-0_21,Stolk2020ValidationOI}. These results show that \tshacl tools, and in particular validators, can benefit areas where technologies other than \tshacl are already established.

\section{Conclusion}

Within this review we examined \tshacl, a constraint language that can be used to validate \trdf graphs. These constraints can be used to describe the properties of a graph, to detect possible errors in the data or provide data quality assurances. In this review we first presented the main concepts of the \tshacl specification, such as the concept of \emph{shapes}, and their two main components, \emph{targets} and \emph{constraints}. We discussed the primary way to perform \tshacl validation, using \tsparql queries, and how the semantics of validation can be abstracted with the concept of \emph{assignments}. 

While the \tshacl specification describes how validation should be performed, its semantics is left implicit and not formally defined. We have extensively discussed studies that address this problem. In particular, we reviewed a complete formalisation of \tshacl into a fragment of first order logic called \SCL . This formalisation lays bare several properties of \tshacl , and provides decidability and complexity results for several \tshacl-related decision problems. 
Another important line of work focuses on defining potential extensions of \tshacl semantics that can be used in the face of \emph{recursion}. The \tshacl specification, in fact, allows constraints to be recursively defined, but it does not define its semantics. We also presented existing work studying the interaction of \tshacl with inference rules. Datasets are often dynamic objects, and several questions arise when considering the effects of this dynamism on the constraints imposed over them.

From the point of view of maturity and level of adoption of the \tshacl technology, we reviewed several implementations of \tshacl validators, which are now integrated in many mainstream \trdf databases, and several tools designed to facilitate the creation and management of \tshacl documents. Several approaches, in particular, provide automated or semi-automated ways of generating suitable \tshacl documents from a diverse range of sources, such as graph data, ontologies, or natural language texts. Existing efforts in mapping other constraint/validation languages into \tshacl is also worth noting, as it suggests that the usefulness of \tshacl could be extended to support other existing technologies. 
While the true extent of \tshacl adoption is hard to establish, since not all usages of \tshacl are publicly documented, we found evidence of its usage in several areas, such as to facilitate data sharing, to validate dataset against policies, and to detect errors in datasets.

Despite the wealth of work on this topic, \tshacl is still a recent specification, and a number of important directions for future work still exist. For example, there are opportunities to optimise \tshacl validators for particular type of constraints, or for particular scenarios, like for highly dynamic databases. More studies are needed to properly assess the usage of \tshacl in practical applications, and what types of constraints are more commonly used and how. While the full semantics of \tshacl has been formally defined, more work is needed to formally establish its relation with other constraint languages. 
It is also important to notice that most of the pieces of work reviewed in this article limit their scope to ad-hoc subsets of the \tshacl specification. In addition to the custom requirements of each application, this is commonly done in order to avoid excessively complex language components. At the same time, it is often difficult to understand what these subsets exactly are as they are not always explicitly defined. Therefore, there might be scope to define reusable fragments of \tshacl , that could fill the role of lightweight but expressive alternatives to the full language, similarly to how OWL fragments are defined. It might also be beneficial, for similar reasons, to converge towards a single standard or ``preferred'' semantics for \tshacl recursion, which could be defined in a future version of the specification.

\bibliographystyle{splncs04}
\bibliography{bibliography}

\begin{thebibliography}{10}
\providecommand{\url}[1]{\texttt{#1}}
\providecommand{\urlprefix}{URL }
\providecommand{\doi}[1]{https://doi.org/#1}

\bibitem{ABBV16}
{A.~Amarilli and M.~Benedikt and P.~Bourhis and M.~Vanden Boom}: {Query
  Answering with Transitive and Linear-Ordered Data.} In: IJCAI'16. pp.
  893--899 (2016)

\bibitem{al2020automated}
Al~Bassit, A., Krasnashchok, K., Skhiri, S., Mustapha, M.: {Automated
  Compliance Checking with SHACL} (2020)

\bibitem{SHACLstableModelSemantics}
Andresel, M., Corman, J., Ortiz, M., Reutter, J.L., Savkovic, O., Simkus, M.:
  {Stable Model Semantics for Recursive SHACL}. In: Proceedings of The Web
  Conference 2020. p. 1570–1580. WWW ’20 (2020)

\bibitem{benedikt2017benchmarking}
Benedikt, M., Konstantinidis, G., Mecca, G., Motik, B., Papotti, P., Santoro,
  D., Tsamoura, E.: Benchmarking the chase. In: Proceedings of the 36th ACM
  SIGMOD-SIGACT-SIGAI Symposium on Principles of Database Systems. pp. 37--52.
  ACM (2017)

\bibitem{Ber66}
Berger, R.: {The Undecidability of the Domino Problem.} MAMS  \textbf{66},
  1--72 (1966)

\bibitem{bonevashapedesigner}
Boneva, I., Dusart, J., Fern{\'a}ndez~Alvarez, D., Gayo, J.E.L.: {Shape
  Designer for ShEx and SHACL Constraints}. {ISWC 2019 - 18th International
  Semantic Web Conference, Poster} (Oct 2019)

\bibitem{RDFS}
Brickley, D., Guha, R.: {RDF} schema {1.1}. {W3C} recommendation, W3C (Feb
  2014), https://www.w3.org/TR/2014/REC-rdf-schema-20140225/

\bibitem{ns2021schemabacked}
{\v{C}}er{\={a}}ns, K., Ov{\v{c}}i{\c{n}}{\c{n}}ikova, J., Boj{\={a}}rs, U.,
  Grasmanis, M., L{\={a}}ce, L., Rom{\={a}}ne, A.: {Schema-Backed Visual
  Queries over Europeana and other Linked Data Resources}. In: ESWC2021 Poster
  and Demo Track (2021)

\bibitem{Ceri1989Datalog}
Ceri, S., Gottlob, G., Tanca, L.: What you always wanted to know about datalog
  (and never dared to ask). IEEE Transactions on Knowledge and Data Engineering
   \textbf{1}(1),  146--166 (1989)

\bibitem{astrea}
Cimmino, A., Fern{\'a}ndez-Izquierdo, A., Garc{\'i}a-Castro, R.: {Astrea:
  Automatic Generation of SHACL Shapes from Ontologies}. In: The Semantic Web.
  pp. 497--513. Springer International Publishing, Cham (2020)

\bibitem{corman2019shacl2sparql}
Corman, J., Florenzano, F., Reutter, J.L., Savkovic, O.: {SHACL2SPARQL:
  Validating a SPARQL Endpoint against Recursive SHACL Constraints}. In:
  International Semantic Web Conference ISWC Satellite Events. pp. 165--168
  (2019)

\bibitem{SHACL2SPARQLtranslation}
Corman, J., Florenzano, F., Reutter, J.L., Savkovi{\'{c}}, O.: {Validating
  SHACL Constraints over a Sparql Endpoint}. In: The Semantic Web -- ISWC 2019.
  pp. 145--163 (2019)

\bibitem{Corman2018SHACL}
Corman, J., Reutter, J.L., Savkovi{\'{c}}, O.: {Semantics and Validation of
  Recursive SHACL}. In: The Semantic Web -- ISWC 2018. pp. 318--336 (2018)

\bibitem{Hayes2014GeneralisedRDF}
Cyganiak, R., Wood, D., Markus~Lanthaler, G.: {RDF 1.1 Concepts and Abstract
  Syntax}. {W3C} {R}ecommendation, {W3C} (2014),
  \url{http://www.w3.org/TR/2014/REC-rdf11-concepts-20140225/}

\bibitem{DK19}
Danielski, D., Kieronski, E.: {Finite Satisfiability of Unary Negation Fragment
  with Transitivity.} In: MFCS'19. pp. 17:1--15. LIPIcs 138, Leibniz-Zentrum
  fuer Informatik (2019)

\bibitem{DiCiccio2019}
Di~Ciccio, C., Ekaputra, F.J., Cecconi, A., Ekelhart, A., Kiesling, E.:
  {Finding Non-compliances with Declarative Process Constraints Through
  Semantic Technologies}. In: Cappiello, C., Ruiz, M. (eds.) Information
  Systems Engineering in Responsible Information Systems. pp. 60--74 (2019)

\bibitem{7936162}
Ekaputra, F.J., Lin, X.: {SHACL4P: SHACL constraints validation within
  Protégé ontology editor}. In: 2016 International Conference on Data and
  Software Engineering (ICoDSE). pp.~1--6 (2016)

\bibitem{TravSHACL2021}
Figuera, M., Rohde, P.D., Vidal, M.E.: {Trav-SHACL: Efficiently Validating
  Networks of SHACL Constraints}. In: Proceedings of the Web Conference 2021.
  p. 3337–3348. WWW '21, Association for Computing Machinery, New York, NY,
  USA (2021)

\bibitem{Gangemi2009}
Gangemi, A., Presutti, V.: {Ontology Design Patterns}, pp. 221--243. Springer
  Berlin Heidelberg, Berlin, Heidelberg (2009)

\bibitem{JLMS18}
Jung, J., Lutz, C., Martel, M., Schneider, T.: {Querying the Unary Negation
  Fragment with Regular Path Expressions.} In: ICDT'18. pp. 15:1--18.
  OpenProceedings.org (2018)

\bibitem{k2019modellingConstruction}
K~Soman, R.: {Modelling construction scheduling constraints using shapes
  constraint language (SHACL)}. In: 2019 European Conference on Computing in
  Construction. pp. 351--358. University College Dublin (2019)

\bibitem{PMID:34042768}
Keuchel, D., Spicher, N.: {Automatic Detection of Metadata Errors in a Registry
  of Clinical Studies Using Shapes Constraint Language (SHACL) Graphs}. Studies
  in health technology and informatics  \textbf{281},  372—376 (May 2021)

\bibitem{2017SHACL}
Knublauch, H., Kontokostas, D.: {Shapes Constraint Language (SHACL)}. {W3C}
  {R}ecommendation, {W3C} (2017), \url{https://www.w3.org/TR/shacl/}

\bibitem{Kontokostas2014RuleValidation}
Kontokostas, D., Westphal, P., Auer, S., Hellmann, S., Lehmann, J.,
  Cornelissen, R., Zaveri, A.: {Test-driven Evaluation of Linked Data Quality}.
  In: Proceedings of the 23rd International Conference on World Wide Web. pp.
  747--758. WWW '14, ACM (2014)

\bibitem{9205800}
Larhrib, M., Escribano, M., Cerrada, C., Escribano, J.J.: {Converting OCL and
  CGMES Rules to SHACL in Smart Grids}. IEEE Access  \textbf{8},
  177255--177266 (2020)

\bibitem{martin2020shapecontainment}
Leinberger, M., Seifer, P., Rienstra, T., L{\"a}mmel, R., Staab, S.: {Deciding
  SHACL Shape Containment through Description Logics Reasoning}. In: The
  Semantic Web -- ISWC 2020. Springer International Publishing (2020),
  (\emph{this volume})

\bibitem{MartinTypeCheckingCodeWithSHACL2019}
Leinberger, M., Seifer, P., Schon, C., L{\"a}mmel, R., Staab, S.: {Type
  Checking Program Code Using SHACL}. In: The Semantic Web -- ISWC 2019. pp.
  399--417. Springer International Publishing, Cham (2019)

\bibitem{liebervisual}
Lieber, S., De~Meester, B., Heyvaert, P., Br{\"u}ckmann, F., Wambacq, R.,
  Mannens, E., Verborgh, R., Dimou, A.: {Visual Notations for Viewing and
  Editing RDF Constraints with UnSHACLed}. Semantic Web  (2021), under review

\bibitem{lieber2020statistics}
Lieber, S., Dimou, A., Verborgh, R.: {Statistics about Data Shape Use in RDF
  Data}. In: ISWC (Demos/Industry) (2020)

\bibitem{marnette2009generalized}
Marnette, B.: Generalized schema-mappings: from termination to tractability.
  In: Proc. of the twenty-eighth ACM SIGMOD-SIGACT-SIGART symp. on Principles
  of database systems. pp. 13--22. ACM (2009)

\bibitem{Protege2015}
Musen, M.A., {Protégé Team}: {The Protégé Project: A Look Back and a Look
  Forward}. AI matters  \textbf{1}(4),  4—12 (June 2015)

\bibitem{8249433}
Nenadić, K.R., Gavrić, M.M., Đurđević, V.I.: {Validation of CIM datasets
  using SHACL}. In: 2017 25th Telecommunication Forum (TELFOR). pp.~1--4 (2017)

\bibitem{omranlearning}
Omran, P.G., Taylor, K., Mendez, S.R., Haller, A.: {Learning SHACL Shapes from
  Knowledge Graphs}. Semantic Web  (2021), \emph{under review}

\bibitem{Pandit2018UsingODP4SHACL}
Pandit, H.J., O’Sullivan, D., Lewis, D.: {Using Ontology Design Patterns To
  Define SHACL Shapes}. In: 9th Workshop on Ontology Design and Patterns
  (WOP2018), International Semantic Web Conference (ISWC). pp. 67--71 (2018)

\bibitem{shacl4gdpr}
Pandit, H.J., O'Sullivan, D., Lewis, D.: {Test-Driven Approach Towards GDPR
  Compliance}. In: Semantic Systems. The Power of AI and Knowledge Graphs. pp.
  19--33. Springer International Publishing, Cham (2019)

\bibitem{pareti2021satisfiability}
Pareti, P., Konstantinidis, G., Mogavero, F.: {Satisfiability and Containment
  of Recursive SHACL} (2021), arXiv preprint 2108.13063

\bibitem{pareti2020}
Pareti, P., Konstantinidis, G., Mogavero, F., Norman, T.J.: {SHACL
  Satisfiability and Containment}. In: The Semantic Web -- ISWC 2020. pp.
  474--493. Springer International Publishing, Cham (2020)

\bibitem{pareti2019c}
Pareti, P., Konstantinidis, G., Norman, T.J., \c{S}ensoy, M.: {SHACL
  Constraints with Inference Rules}. In: The Semantic Web -- ISWC 2019.
  Springer International Publishing (2019)

\bibitem{Rob71}
Robinson, R.: {Undecidability and Nonperiodicity for Tilings of the Plane.} IM
  \textbf{12},  177--209 (1971)

\bibitem{10.1007/978-3-030-21348-0_21}
Savkovi{\'{c}}, O., Kharlamov, E., Lamparter, S.: {Validation of SHACL
  Constraints over KGs with OWL 2 QL Ontologies via Rewriting}. In: The
  Semantic Web. pp. 314--329. Springer International Publishing, Cham (2019)

\bibitem{SHACLbenchmark2020}
Schaffenrath, R., Proksch, D., Kopp, M., Albasini, I., Panasiuk, O., Fensel,
  A.: {Benchmark for Performance Evaluation of SHACL Implementations in Graph
  Databases}. In: Rules and Reasoning. pp. 82--96. Springer International
  Publishing, Cham (2020)

\bibitem{SeaborneSPARQL2013}
Seaborne, A., Harris, S.: {SPARQL} 1.1 query language. {W3C} recommendation,
  W3C (Mar 2013), https://www.w3.org/TR/2013/REC-sparql11-query-20130321/

\bibitem{10.1007/978-3-030-35646-0_9}
{\v{S}}enk{\'y}{\v{r}}, D.: {SHACL Shapes Generation from Textual Documents}.
  In: Enterprise and Organizational Modeling and Simulation. pp. 121--130.
  Springer International Publishing, Cham (2019)

\bibitem{spahiu2018towards}
Spahiu, B., Maurino, A., Palmonari, M.: {Towards Improving the Quality of
  Knowledge Graphs with Data-driven Ontology Patterns and SHACL.} In:
  International Semantic Web Conference ISWC Sattelite Events. pp. 103--117
  (2018)

\bibitem{ABSTAT2016}
Spahiu, B., Porrini, R., Palmonari, M., Rula, A., Maurino, A.: {ABSTAT:
  Ontology-Driven Linked Data Summaries with Pattern Minimalization}. In: The
  Semantic Web. pp. 381--395. Springer International Publishing, Cham (2016)

\bibitem{stani2019design}
Stani, E.: {Design reusable SHACL shapes and implement a linked data validation
  pipeline}. Code4Lib Journal  \textbf{45} (2019)

\bibitem{Stolk2020ValidationOI}
Stolk, S., McGlinn, K.: {Validation of IfcOWL datasets using SHACL}. In:
  Proceedings of the 8th Linked Data in Architecture and Construction Workshop.
  pp. 91--104 (2020)

\bibitem{ShEX2019}
Thornton, K., Solbrig, H., Stupp, G.S., Labra~Gayo, J.E., Mietchen, D.,
  Prud'hommeaux, E., Waagmeester, A.: {Using Shape Expressions (ShEx) to Share
  RDF Data Models and to Guide Curation with Rigorous Validation}. In: The
  Semantic Web. pp. 606--620. Springer International Publishing, Cham (2019)

\bibitem{OWL2}
{W3C OWL Working Group}: {OWL 2 Web Ontology Language Document Overview (Second
  Edition)}. {W3C} recommendation, W3C (Dec 2012),
  https://www.w3.org/TR/2012/REC-owl2-overview-20121211/

\bibitem{Wan61}
Wang, H.: {Proving Theorems by Pattern Recognition II.} BSTJ  \textbf{40},
  1--41 (1961)

\bibitem{Schimatos2020}
Wright, J., Rodr{\'i}guez~M{\'e}ndez, S.J., Haller, A., Taylor, K., Omran,
  P.G.: {Sch{\'i}matos: A SHACL-Based Web-Form Generator for Knowledge Graph
  Editing}. In: The Semantic Web -- ISWC 2020. pp. 65--80 (2020)

\end{thebibliography}

\end{document}